\documentclass[fleqn,10pt]{wlscirep}

\usepackage[utf8]{inputenc}
\usepackage[T1]{fontenc}
\usepackage{graphicx}
\usepackage{float}
\usepackage{subcaption}
\usepackage[normalem]{ulem}
\usepackage{wrapfig}
\usepackage{caption}
\usepackage{algorithm}
\usepackage{algpseudocode}
\usepackage{float}
\usepackage{makecell}
\usepackage[dvipsnames]{xcolor}
\usepackage[table]{xcolor}
\usepackage{multirow}
\usepackage{silence}

\hypersetup{hidelinks}

\newcommand{\best}[1]{\textbf{#1}}
\newcommand{\second}[1]{\uline{#1}}
\newcommand{\pairwin}[1]{\colorbox{gray!12}{\strut #1}}

\title{Toxicity Assessment in Preclinical Histopathology via Class-Aware Mahalanobis Distance for Known and Novel Anomalies}

\author[1,*]{Olga Graf}
\author[2]{Dhrupal Patel}
\author[3]{Peter Groß}
\author[3]{Charlotte Lempp}
\author[1]{Matthias Hein}
\author[4]{Fabian Heinemann}
\affil[1]{Tübingen AI Center, University of Tübingen, Tübingen, Germany}
\affil[2]{Toxicologic Pathologist, Gujarat, India}
\affil[3]{Boehringer Ingelheim Pharma GmbH and Co., Biberach an der Riß, Germany}
\affil[4]{Boehringer Ingelheim GmbH, Biberach an der Riß, Germany}

\affil[*]{Corresponding author: olga.graf@uni-tuebingen.de}


\begin{abstract}

Drug-induced toxicity is a leading cause of preclinical and early-clinical failure, making early detection critical. Histopathology is the gold standard for toxicity assessment but relies on expert pathologists, creating a bottleneck for large-scale screening. We introduce an AI-based anomaly detection framework for whole-slide images (WSIs) of rodent liver that identifies healthy tissue and known pathologies (anomalies) and flags samples without training data as out-of-distribution (OOD). We evaluate OOD detection on two held-out categories: apoptosis (single-cell, near-OOD) and staining/processing artifacts (heterogeneous, far-OOD). We build a novel pixelwise-annotated dataset and fine-tune a pre-trained Vision Transformer (DINOv2) via Low-Rank Adaptation (LoRA) for segmentation, then use the Mahalanobis distance for OOD detection with class-specific thresholds. Optimizing the false positive rate subject to a predefined constraint on the false negative rate yields only 0.16\% of pathological tissue classified as healthy and 0.35\% of healthy tissue classified as pathological. Our false negative rate does not penalise cross-type errors, reflecting the safety-first objective of never overlooking a lesion; under the stricter correct-class criterion our method assigns 93.93\% of ID and 89.38\% of OOD findings to their own class. The study demonstrates technical feasibility of pixel-level anomaly detection for mouse liver histopathology, indicating possible applications in improving preclinical workflows and drug development efficiency. 
\end{abstract}

\begin{document}

\flushbottom
\maketitle
\thispagestyle{empty}

\section*{Introduction}

Drug development is a costly and time intensive process, with declining productivity observed over several decades. This trend is encapsulated by Eroom's Law, which observes that the inflation-adjusted cost of developing a new medicine roughly doubles every nine years despite technological advances \cite{scannell2012diagnosing}. This trajectory shows the need for strategies that reduce late-stage attrition (failure of drug candidates).

Toxicity is a major cause of preclinical and Phase 1 failures \cite{waring2015analysis}, making early and more accurate toxicity detection critical to preventing costly late-stage attrition and allowing more cost-effective drug development. In practice, such toxicity assessment is largely performed during preclinical \textit{in vivo} studies, typically conducted in rodents. While ongoing efforts aim to reduce, refine, and replace animal testing (3R) \cite{everitt2015future}, such studies remain essential for capturing whole-organism complexity and for assessing both efficacy and toxicological effects prior to human trials. Analysis of such \textit{in vivo} efficacy and toxicology studies typically involves histopathologic analysis, which inherently depends on expert pathologists whose limited availability creates a significant bottleneck for large-scale toxicity screening \cite{zynger2019understanding}. Recent advances in computational pathology and artificial intelligence are increasingly alleviating this constraint by enabling comprehensive, quantitative analysis of whole-slide images (WSIs) \cite{heinemann2022deep, campanella2019clinical}.

In this work, we propose an AI-based anomaly detection (AD) system which can be employed in efficacy studies and exploratory toxicology studies. We envision two complementary applications:
\begin{itemize}
    \item \emph{Secondary safety layer in exploratory toxicology studies}, which occur in many organizations before the formal non-Good Laboratory Practice (GLP)-regulated toxicology studies. AD can run alongside the pathologist’s review, highlighting regions of interest, reducing the risk of missing subtle tissue alterations, and improving overall workflow efficiency.
    \item \emph{Early toxicity screening during efficacy studies}, where tissue is collected for efficacy evaluation. Non-target tissues (e.g., liver, heart, spleen) can be screened for early toxicity signals,  enabling earlier no-go decisions and reducing late stage attrition. Although doses are often low, early microscopic alterations may already occur. Thus, automated screening can provide early mechanistic insight and lead to mitigation measures, which improve the quality of a pharma pipeline\cite{thoolen2010proliferative, ying2006modern}.
\end{itemize}

We note that these applications are not directly tested in this study, however, they motivate our system design and evaluation.

\begin{figure}[H]
\centering
\includegraphics[width=1\linewidth]{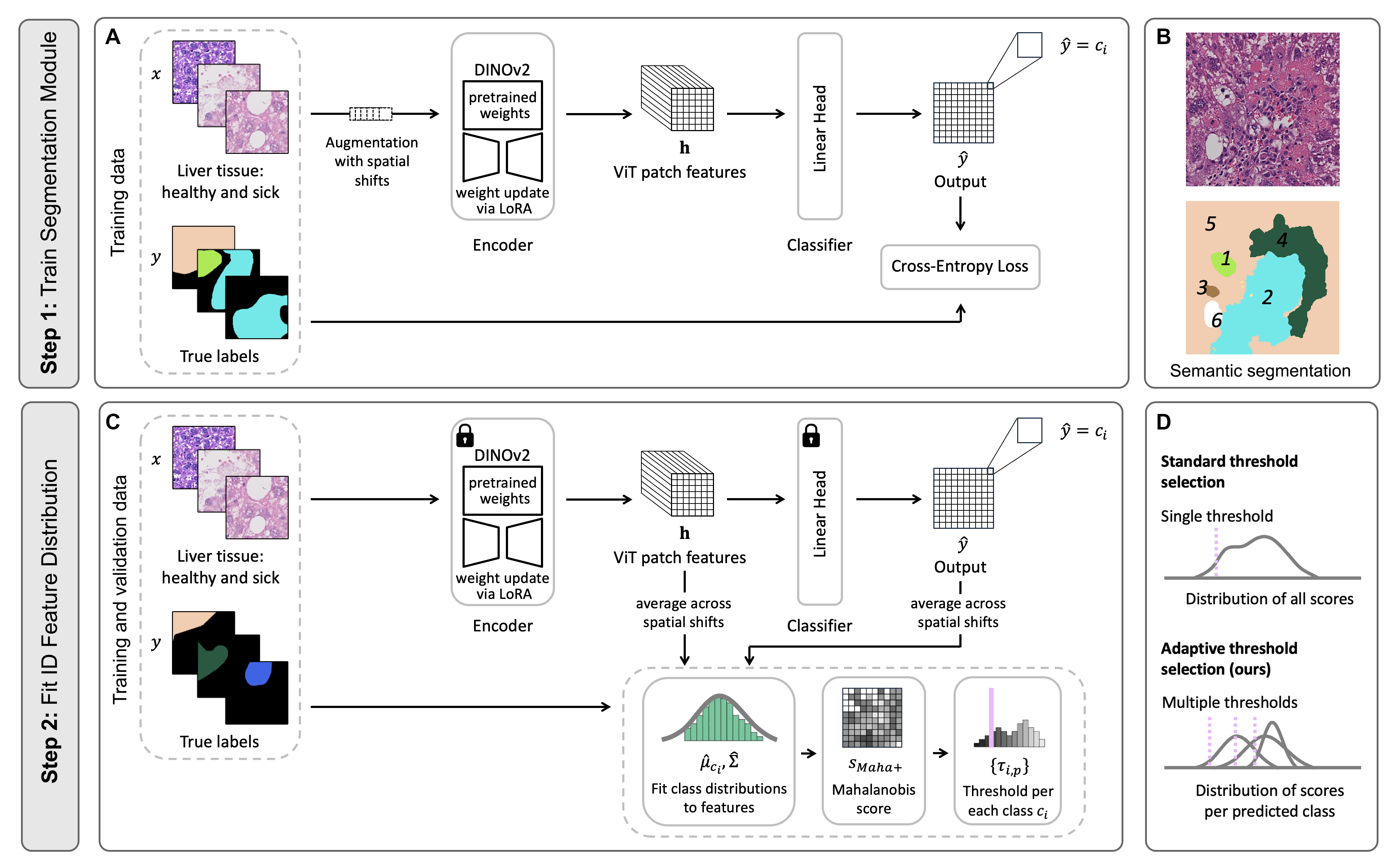}
\caption{Two-step training approach of the system that performs semantic segmentation of tissue states with available training data (in-distribution (ID) data; i.e., healthy tissue and a set of common pathologies); additionally, it can detect tissue states without training examples as out-of-distribution (OOD) anomalies (i.e., rare pathologies, or artifacts). \textbf{A}. In Step 1, the system is trained as a conventional segmentation approach using a foundation model (DINOv2), adjusted to the known histopathology data. \textbf{B}. Example of a segmentation output on mouse liver stained in H\&E. On this patch, alongside with healthy tissue (5), several anomaly types are detected: ballooning (1), inflammation (2), mitosis (3), and necrosis (4). Segmentation also shows a void in the tissue (6), resulting from cross-sectioning the vascular system. \textbf{C}. In Step 2, the trained segmentation model is used to encode images pixelwise into descriptive features, which allow to detect anomalies. For each ID histopathology class, the distribution in the feature space is obtained and the mean (class-wise) and covariance matrix (shared over classes) of these features is used to obtain  the Mahalanobis score. We determine thresholds on the scores per class, motivated by a strong class-specific variability. At inference time we use the pixelwise scores of all classes to discriminate between known ID data and unknown OOD data using the encoded features of a test image. \textbf{D}. Schematic difference between the standard threshold selection strategy where all scores from the training/validation set form a single distribution and adaptive per-class selection strategy where we decompose the global score distribution into class-specific distributions and assign individual thresholds.}
\label{fig:main_scheme}
\end{figure}

From a machine learning perspective, the identification of unexpected or abnormal tissue alterations can be formulated as a distributional shift detection problem. Such problem setting is addressed in the literature as \emph{out-of-distribution} (OOD) \emph{detection}, with methods ranging from classical one-class classifiers such as the one-class SVM\cite{scholkopf2001}, which estimates a compact support for \emph{in-distribution} (ID) data, to a variety of neural post-processing techniques. MSP\cite{hendrycks2017} uses the maximum softmax probability as a confidence-based OOD score, while Max-Logit\cite{hendrycks2022} applies the same idea directly to logits. Energy-based detection\cite{liu2020energy} instead leverages the log-sum-exp (energy) of logits. KL-Matching\cite{hendrycks2022} compares a test-time probability vector to class-wise mean probability profiles via Kullback-Leibler (KL) divergence. The Mahalanobis distance-based method\cite{mahalanobis1936proc,lee2018} models class-conditional Gaussian feature distributions and uses distance to the nearest class mean as the OOD score. ReAct\cite{sun2021react} improves robustness by truncating feature activations above a learned threshold. All the aforementioned methods do not require OOD data at training/calibration time, unlike, e.g., ODIN\cite{liang2018odin} or Outlier Exposure\cite{hendrycks2019oe}, which is an important practical requirement in histopathology setting. Moreover, they are computationally feasible for semantic segmentation on very large WSIs, while KNN-based methods\cite{sun2022dnn} or ensemble methods\cite{vyas2018leaveout} are not feasible due to GPU memory limitations. Reconstruction-based methods (Autoencoders (AE)\cite{bergmann2019ssim, gong2019memae}, Variational Autoencoders (VAE) \cite{kingma2014vae,sato2019uncertainty,liu2020explainvae}, Generative Adversarial Network (GAN)\cite{sabokrou2018alocc,pidhorskyi2018gpnd,akcay2018ganomaly}, or diffusion model (DM)\cite{liu2025diffusion_survey}-based reconstruction) suffer from drawbacks, such as being prone to overgeneralizing to OOD patterns and thus reducing anomaly scores \cite{perera2019ocgan}, challenges to train these approaches, and computational cost at inference.

OOD detection in the context of semantic image segmentation is still in its early stages, with a limited number of works addressing this problem. The most widely considered application is autonomous driving\cite{Zhao2024semseg, tian2022energyabstention, liu2023rpl}; in the context of medical imaging first works show results for MRI data\cite{karimi2023segmentation} and histopathology\cite{dippel2024}. However, these works can only differentiate between ID and OOD data, but not between ID classes, and known anomaly types are not used during training. Moreover, these methods yield separated segmentation and anomaly maps.

In the context of histopathology, however, data for several anomaly types is typically available and can improve a detection system if used for training. Moreover, a single interpretable map is more suitable in a practical setting, as it aligns with the way pathologists reason. A promising approach to provide such a unified segmentation and detection by Wang et al.\cite{wang2025} employs a U-Net architecture and demonstrates results on hyperspectral images from publicly accessible surgical datasets.

To the best of our knowledge, the approach by Zingman et al.\cite{zingman2024learning} is the only AD method developed for toxicity screening. It uses healthy data for training and combines a convolutional neural network (CNN) backbone with a one-class SVM. While this method demonstrates the feasibility of automated tox screening, its tile-wise formulation and binary "healthy" vs. "anomaly" classification limits spatial precision and interpretability.

\begin{figure}[H]
\centering
\includegraphics[width=0.75\textwidth]{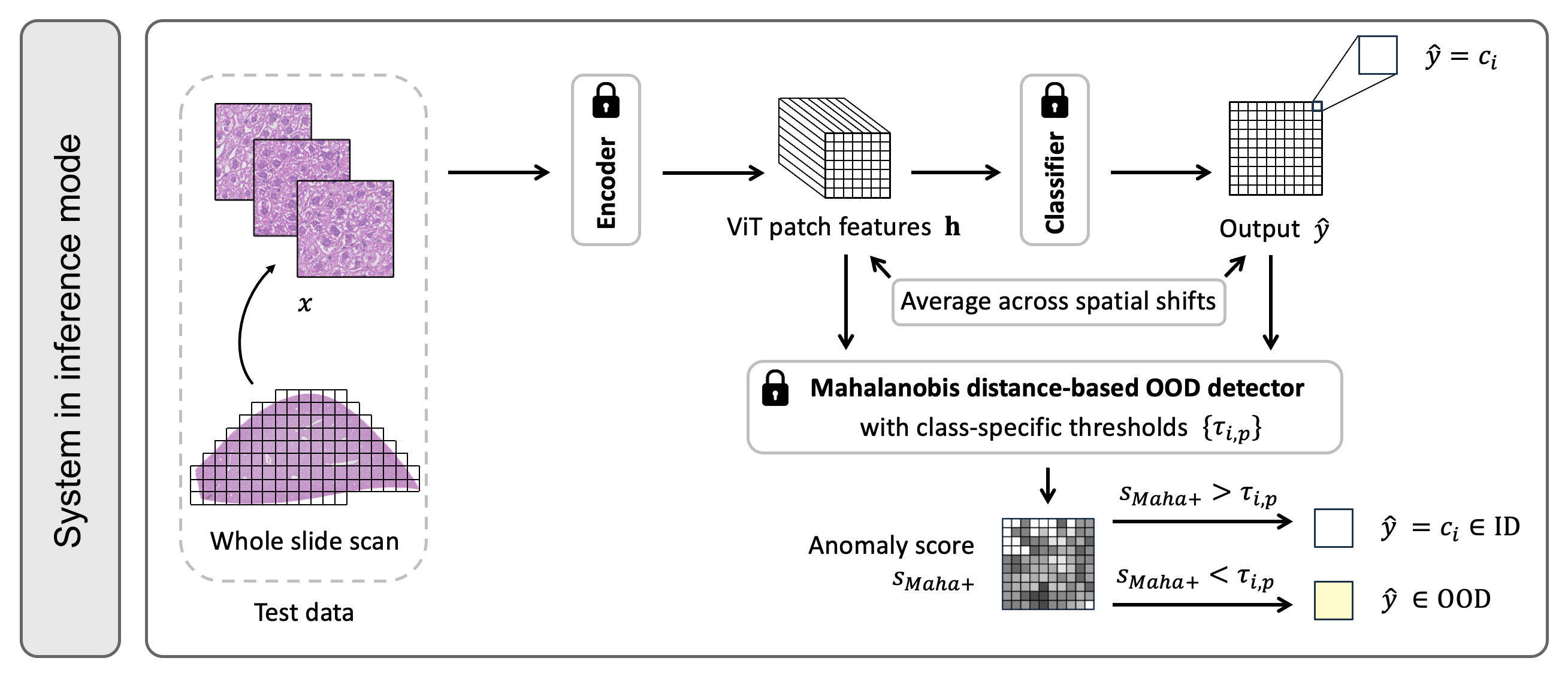}
    \captionof{figure}{Inference workflow for segmentation and anomaly detection. WSIs (H\&E-stained mouse liver) are divided into patches and processed by an encoder to extract spatially resolved feature representations ($\mathbf{h}$). These features are passed to a classifier to predict the most likely class ($c_i$), and the outputs are averaged across spatial shifts for increased robustness. For each predicted class, the Mahalanobis distance ($s_{Maha+}$) between the features and the estimated class mean is computed using the estimated covariance matrix. Based on the class-specific threshold ($\tau_{i,p}$), a spatial location in a patch is either confirmed as in-distribution (ID) with class $c_i$ or flagged as out-of-distribution (OOD) when the score is below the threshold.}
    \label{fig:inference}    
\end{figure}

Here, we introduce a pixelwise AD framework for histology that unifies (i) semantic segmentation of healthy tissue and common pathology types using available training data and (ii) detection of rare pathologies without training data, validated on two held-out categories: apoptosis and staining/processing artifacts. This mirrors typical laboratory conditions: data is abundant for healthy tissue and for common pathologies, but little to no data is available for rare pathologies.

Our system builds on deep image representations of a pre-trained DINOv2\cite{oquab2023dinov2} foundation model, adapted for semantic segmentation of our histological data using Low-Rank Adaptation (LoRA) \cite{hu2022lora}. Our final system provides pixelwise predictions for healthy and known anomaly classes. In order to be able to flag unknown anomalies as OOD, we extend the Mahalanobis distance-based method of Lee et al.~\cite{lee2018}. As the Mahalanobis scores are strongly varying between healthy and known anomalies, we propose to use class-wise thresholds, which improve joint segmentation and anomaly detection performance.

We validate our approach in rodent liver WSIs corresponding to a study with known histopathological findings. We show that the system detects and quantifies multiple categories of anomalies associated with toxicological effects. By delivering granular, pathology-resolved outputs for known anomalies, the method provides richer context for pathologists and earlier signal detection in efficacy studies. These capabilities could help mitigate productivity pressures by reducing costly late stage failures through earlier, data-driven decision making in preclinical drug development.

\section*{Results and Discussion}

\subsection*{Overview of the unified segmentation and anomaly detection approach}

Our dataset includes WSIs of mouse liver with pixelwise annotations for healthy tissue and multiple common pathology types, which we collectively denote as in-distribution (ID) classes. WSIs may also contain previously unseen, rare pathologies for which no training data exists; we refer to these as out-of-distribution (OOD) anomalies. In this work, we will use the collective term \emph{anomalies} for both known (ID) and unknown (OOD) pathology types. The ultimate goal of an AD system is to minimize missed anomalies in line with the safety-first principle, i.e., achieving \emph{false negative rate} (FNR) $\approx 0$, while low \emph {false positive rate} (FPR) remains a secondary goal to limit unnecessary pathologist workload.

Our method (\emph{Maha+} under \emph{Adaptive threshold selection} in Table \ref{tab:ood_results_combined}) follows a two-step training process (Fig.~\ref{fig:main_scheme}). In Step 1, a semantic segmentation model is trained using pixelwise annotations of healthy and anomalous classes. A pretrained DINOv2 foundation model provides expressive representations, which are adapted to our data via LoRA. In Step 2, the anomaly detector is calibrated on training and validation data. Using the segmentation model as an encoder, we introduce a class-conditional anomaly calibration strategy. First, we estimate per-class feature distributions (means and covariances). Additionally, we propose to average features and softmax predictions across spatial shifts to reduce tile-boundary inconsistencies in WSIs. Class-specific Mahalanobis thresholds are then derived to distinguish ID from OOD regions.

At inference time (Fig.~\ref{fig:inference}), the system comprises a \textit{tissue segmentation} module (encoder + classifier) and an \textit{OOD anomaly detection} module. WSIs are tiled, and each tile is processed to obtain pixelwise class predictions and encoder features, again averaged across spatial shifts. Each pixelwise feature vector receives an anomaly score via the Mahalanobis metric, which is compared against the class-specific threshold to flag unknown anomalies.

\subsection*{Known anomaly classification}

\subsubsection*{Leveraging foundation models for semantic segmentation}

Accurate histopathology segmentation requires a backbone capable of producing semantically rich feature vectors that capture complex tissue morphology and distinguish normal from abnormal patterns. ViT-based foundation models have recently demonstrated strong performance in semantic segmentation due to their ability to model global context, including in histopathology \cite{chen2024towards, vorontsov2024foundation}. Our backbone, DINOv2, follows the standard ViT design: images are split into patches, linearly embedded, and processed by transformer layers with multi-head self-attention and feed-forward blocks.

Although DINOv2 provides powerful representations, it lacks dataset-specific label information. We therefore use the frozen DINOv2 encoder \( g_{\phi} \) to extract features \( \mathbf{h}=g_{\phi}(\mathbf{x}) \) and attach a linear segmentation head \( f_{\theta}(\mathbf{h}) = W\mathbf{h} + \mathbf{b} \). To specialize the pretrained features without full fine-tuning, we apply LoRA, inserting low-rank matrices \( A \) and \( B \) into the attention weights: $W' = W + \Delta W = W + BA.$ This strategy allows for an efficient adaptation of large ViTs at low computational cost.  Due to the small number of its parameters, LoRA has a low risk of overfitting on our relatively small, domain-specific dataset. Only the LoRA parameters and the segmentation head \( \theta=\{W,\mathbf{b}\} \) are optimized using labeled data via cross-entropy loss. The fine-tuning workflow is illustrated in Figure~\ref{fig:main_scheme}A.

\noindent
\begin{minipage}[t]{0.36\textwidth}
    \centering
    \vspace{0.1pt}
    \includegraphics[width=0.95\textwidth]{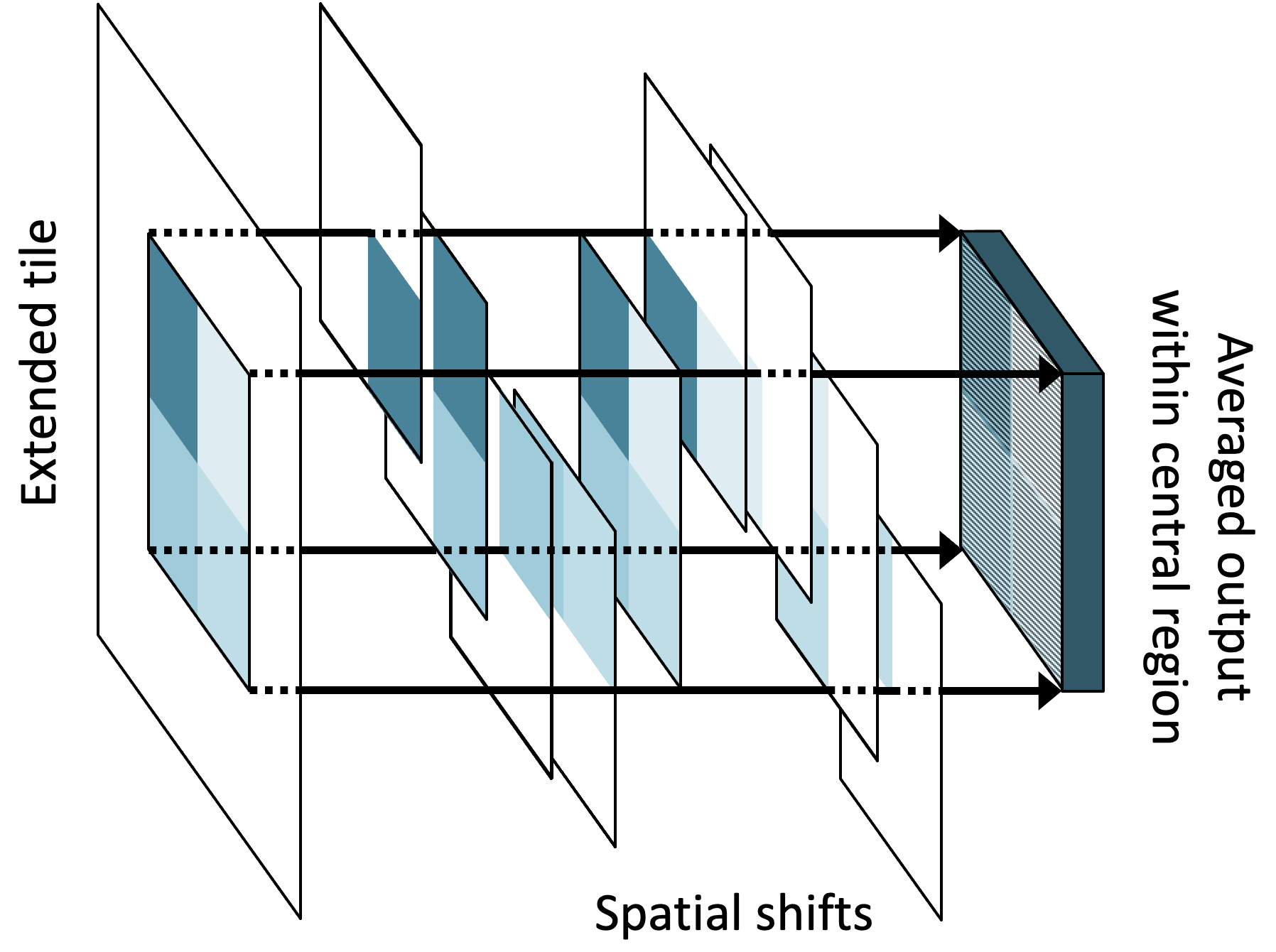}
    \captionsetup{hypcap=false}
    \captionof{figure}{Schematic representation of spatial averaging procedure applied to improve robustness during inference and mitigate boundary artifacts.}
    \captionsetup{hypcap=true}
    \label{fig:shifts}
\end{minipage}
\begin{minipage}[t]{0.635\textwidth}
    \setlength{\parindent}{2em}
    \vspace{0.1pt}

 \subsubsection*{Augmentation and averaging using spatial shifts}

    To improve robustness and generalization, we apply spatial shifting strategy. Unlike in classical semantic segmentation tasks that operate on a full image, WSIs must be divided into smaller tiles for computational feasibility. Applying spatial shifts to these tiles allows the network to see overlapping regions with varying spatial context during training, effectively increasing the amount of training data and promoting learning of more context-aware features. This strategy also reduces sensitivity to patch boundaries during inference. We implement spatial shifting via offset-based tiling (see Appendix A3). Each whole-slide image is divided into \emph{extended} $T \times T$ tiles whose central $t \times t$ ($t < T$) regions form a non-overlapping grid, while overlapping context preserves spatial continuity.
   
    As \textit{augmentation during training}, random spatial shifts are applied by cropping a randomly positioned $t \times t$ subregion from each tile. Because annotations are partial, some crops may fall outside labelled areas; therefore, we sample only shifts containing a class-dependent minimum fraction of annotated pixels (see the Methods section for details).
\end{minipage}

\vspace{2.5pt}
As an additional \textit{inference-time refinement}, we improve WSI-level robustness by mitigating tiling-induced boundary artifacts. Each extended $T \times T$ tile is evaluated under multiple spatial shifts using a $t \times t$ sliding window. For every shift, a $t \times t$ sub-tile is passed through the network, and the final prediction for the central region is obtained by averaging all corresponding softmax outputs (Fig.~\ref{fig:shifts}). This test-time ensemble smoothes boundary effects and improves whole-slide segmentation accuracy.

\subsection*{Unknown anomaly detection}

\begin{minipage}[t]{0.6\textwidth}
    \noindent
    We frame the task of detecting previously unseen anomalies as out-of-distribution detection. Unlike in the standard OOD detection notation, where OOD samples are labeled as positive and ID samples are labeled as negative, in this work we treat \emph{both} ID and OOD anomalies as positive and the healthy class as negative.
    
    \setlength{\parindent}{2em}
    We adopt the following notation:
    \begin{center}
    \(\begin{aligned}
        P &= \text{anomalous tissue samples,}\\
        N &= \text{healthy tissue samples.}
    \end{aligned}\)
    \end{center}
    Therefore, \textit{true positives} in our context are agnostic to specific anomaly type and include anomalies misclassified as another anomalies (e.g., known misclassified as unknown or vice versa, as well as known misclassified as another known). The motivation for deviating from the standard definition is practical: in toxicologic pathology, any abnormal tissue must be flagged, regardless of whether the anomaly type was present in the training data. The distinction between known and unknown anomalies is secondary to the primary safety objective of detecting all pathological deviations.
    \vspace{0.2em}
\end{minipage}\hfill
\begin{minipage}[t]{0.36\textwidth}
    \centering
    \vspace{0.1pt}
    \includegraphics[width=1\textwidth]{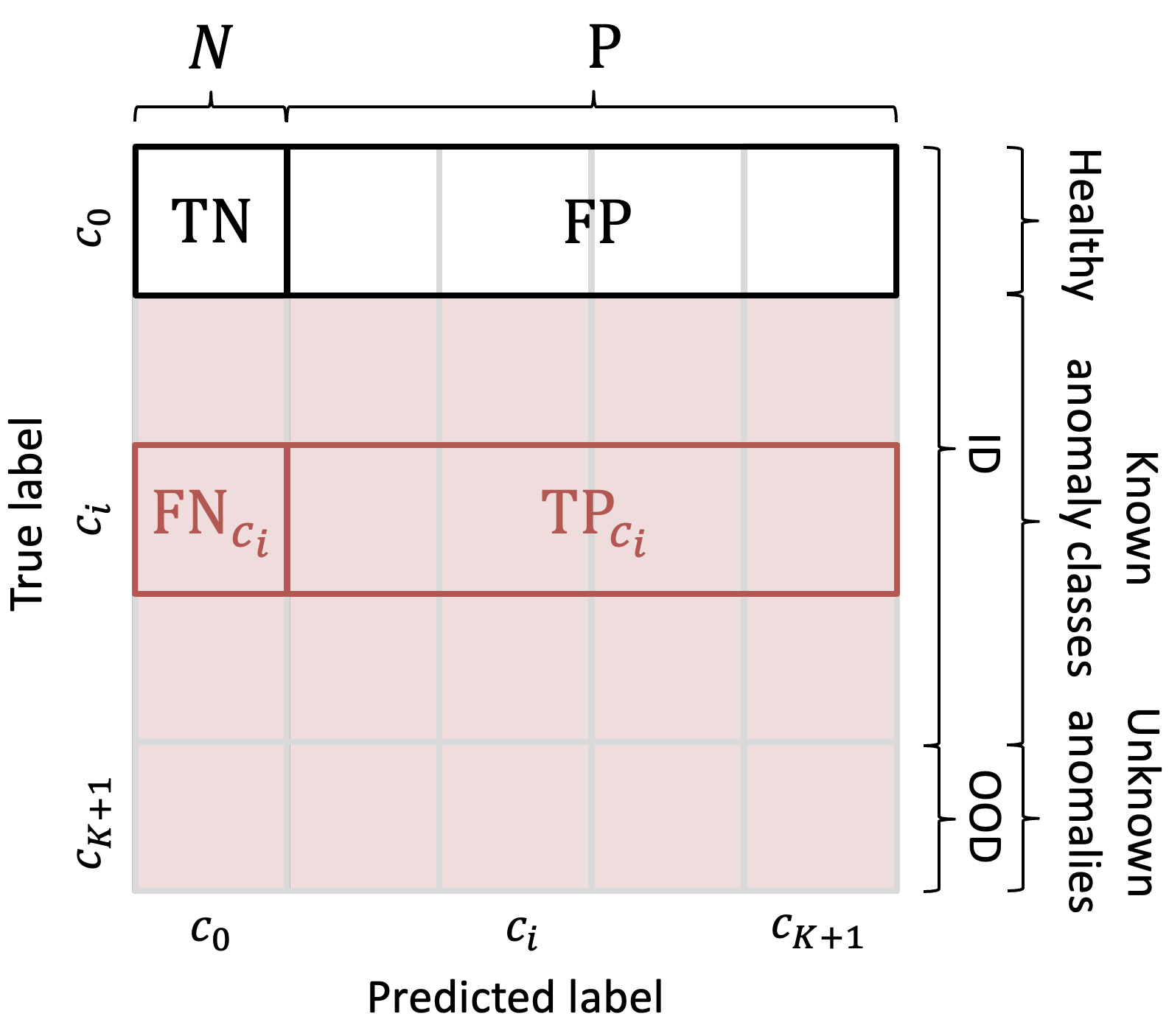}
    \captionsetup{hypcap=false}
    \captionof{figure}{Extended confusion matrix for unified segmentation and anomaly detection.}
    \captionsetup{hypcap=true}
    \label{fig:cm}
\end{minipage}

To evaluate the effectiveness of anomaly detection -- both known and unknown -- we use FNR and FPR. They allow to assess the percentage of all missed anomalies and false alarms, respectively, which is of foremost importance to practitioners, as well as makes the results comparable with binary AD systems. In addition, we perform fine-grained analysis which captures how well the system can differentiate between various anomaly types. For each of three categories -- healthy, ID anomalies and OOD anomalies -- we compute the percentage of \textit{misclassified} samples, which encompasses \textit{misclassification as (another) ID anomaly}, \textit{misclassification as OOD anomaly} (if applicable), and \textit{misclassification as healthy} (if applicable). These numbers can be computed in a straightforward way from the confusion matrix.

We define all metrics using an extended confusion matrix which is shown in Figure \ref{fig:cm}. For this matrix, we define the set of all classes,
\begin{center}
\(
\mathcal{C}=\{c_0:=c_{\text{healthy}},\hspace{0.1cm} \{c_i\}_{i=1}^{K}, \hspace{0.1cm} c_{K+1}:=c_{\text{OOD}}\},
\)
\end{center}
which in addition to the healthy class and $K$ known anomaly classes includes the joint OOD class for all unseen anomalies which are samples labeled as out-of-distribution. For all $K+1$ anomaly classes we define
\begin{center}
\(\begin{aligned}
TP_{c_i} &= \{\, x \mid y = c_i,\ \hat{y} \in \mathcal{C} \setminus  c_{\text{healthy}}  \,\}:
\text{samples of anomaly class $c_i$ predicted as any of the classes except healthy,}
\\
FN_{c_i} &= \{\, x \mid y = c_i,\ \hat{y} = c_{\text{healthy}} \}:
\text{samples of anomaly class $c_i$ wrongly predicted as healthy,}
\\
FP &= \{\, x \mid y = c_{\text{healthy}},\ \hat{y} \in \mathcal{C} \setminus  c_{\text{healthy}}  \, \}: 
\text{healthy samples wrongly predicted as anomaly,}
\\
TN &= \{\, x \mid y = c_{\text{healthy}},\ \hat{y} = c_{\text{healthy}}  \, \}: 
\text{correctly predicted healthy samples,}
\end{aligned}\)
\end{center}
\noindent and the combined measures
\begin{center}
\(\begin{aligned}
\overline{\text{FNR}} = 
\frac{1}{K+1} \sum_{i=1}^{K+1}
\frac{FN_{c_i}}{TP_{c_i} + FN_{c_i}},
\qquad
\text{FPR} = 
\frac{FP}{TN + FP}.
\end{aligned}\)
\end{center}
Note that for $\overline{\text{FNR}}$, individual rates are computed independently for each class 
and then averaged across classes. All classes are therefore treated uniformly, regardless of their prevalence in the dataset. This is justified for histopathological data where certain anomalies may occur much more rarely than the others. Averaging the FNR makes sure that frequently occurring and easy-to-detect anomaly classes do not obscure the contribution of rare and challenging ones.

For each input sample $x$, OOD detection methods provide an \textit{anomaly score} $s(x) \in \mathbb{R}$. We will use the convention where lower values of $s(x)$ indicate a higher likelihood of the sample being out-of-distribution. Given the anomaly score, the final decision is obtained by comparing it to a threshold $\tau$. Samples with scores below the threshold, $s(x) < \tau$, are predicted as OOD, while the rest are predicted as ID. Following a standard strategy in the OOD detection literature, the threshold is determined using true ID samples from the training and/or validation set by fixing the percentage of predicted ID samples to a value $p$.

\begin{figure}[H]
\centering
\includegraphics[width=0.75\linewidth]{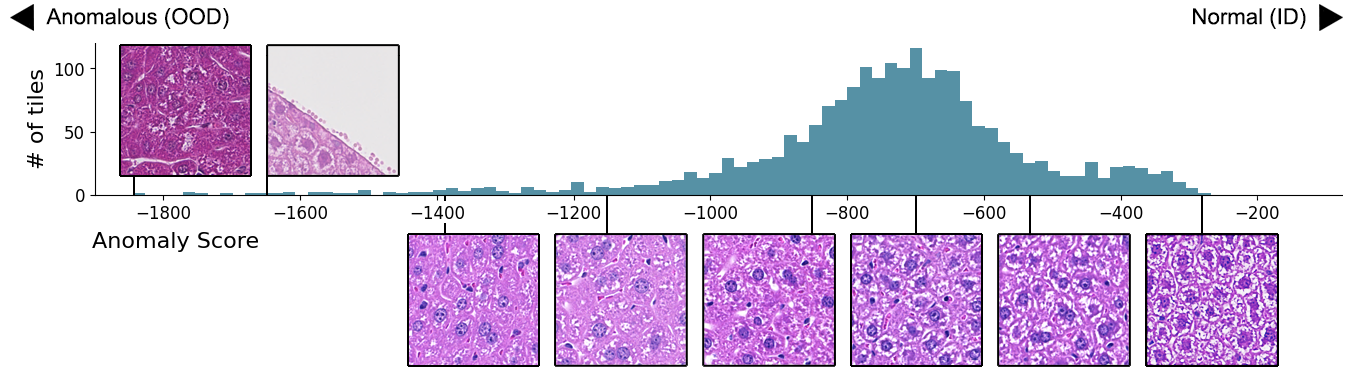}
\caption{Distribution of the mean anomaly score per tile for healthy (no damage model was induced) tissue samples. The samples with the highest scores are the most normal. The outliers with lower scores (i.e., rated as more anomalous) are samples with low glycogen storage as well as samples at the tissue boundary. The samples lying within the central 99\% quantile interval exhibit the gradient transition from normal to low glycogen storage.}
\label{fig:normal_examples}
\end{figure}

Figure \ref{fig:normal_examples} shows the distribution of the mean anomaly scores per tile provided by our method (\emph{Maha+}) for the healthy mouse liver. The selected tiles illustrate how samples from the healthy class with increasing anomaly scores differ from each other. Tiles with higher scores show intact hepatocellular structure, including evenly distributed hepatocyte plates and cytoplasmic clarity consistent with a normal degree of glycogen storage. Among outliers are the samples with lower amounts of glycogen storage, which appear denser and more eosinophilic (i.e., exhibiting stronger pink staining in H\&E slides), as well as samples located at the tissue boundary with collapsed or compressed hepatocytes. While none of these outliers reflect true pathological alterations (lesions), their visual characteristics nonetheless diverge from the typical appearance of the healthy training set, and the variation in the anomaly scores reflects these underlying semantic differences.

\subsubsection*{Mahalanobis distance-based OOD detector}

We adopt the Mahalanobis distance-based OOD detector of Lee et al.~\cite{lee2018}, which models the feature distribution of each class as a multivariate Gaussian and measures the distance of a test feature to these class distributions. This method has been shown to be one of the most promising post-hoc approaches\cite{bitterwolf2023}, and our experiments further corroborate this observation.

To adapt the method to our segmentation setting, we introduce two modifications.  
First, following Müller et al.~\cite{mueller2025}, we compute class statistics using $\ell_2$-normalized features to reduce norm-related variability in ViT representations.  
Second, instead of taking the minimum distance over all classes, we compute the distance only to the predicted class, which produces class-specific score sets compatible with our adaptive threshold selection. Additionally, feature embeddings undergo an averaging procedure across spatial shifts (see Appendix A3 for details), analogous to the prediction averaging described earlier. In Table \ref{tab:ood_results_combined}, the method which includes the proposed modifications is labeled as \emph{Maha+}. For the full mathematical formulation, please refer to Appendix A6.

\begin{figure}[H]
\centering
\includegraphics[width=1\linewidth]{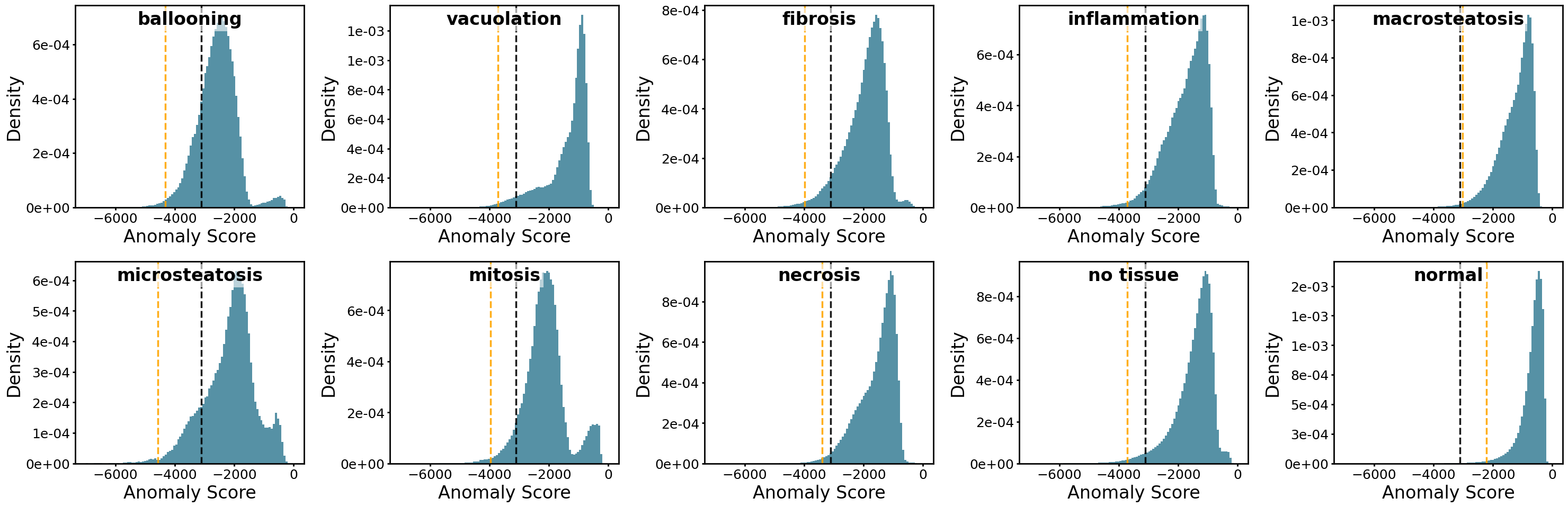}
\caption{Anomaly score histograms for different anomaly classes and the normal (healthy tissue) class, calculated on combined training and validation set. Using a fixed threshold (black dashed line) for all classes to determine OOD samples would lead to many errors for a lot of the ID anomalies. The fixed threshold is set based on accepting 99\% of scores in the combined (all-class) score set. The adaptive threshold (orange dashed line) mitigates this problem, as it is set based on accepting 99\% of scores in each predicted class.}
\label{fig:example}
\end{figure}

\subsubsection*{Adaptive threshold selection}

A key methodological contribution of our work lies in the refinement of threshold selection. Our approach can be applied together with any post-hoc OOD detection method which provides a per-pixel anomaly score.

The standard approach to threshold selection assumes that all known classes have similar anomaly score distributions and therefore considers a single global distribution of anomaly scores of all classes. However, for histological data, this assumption does not hold. Figure \ref{fig:example} shows per-class scores produced by our Mahalanobis distance-based OOD detector (\emph{Maha+}).
We observe that score histograms of anomaly classes differ significantly from the healthy class as well as between each other. The standard threshold choice, based on accepting 99\% of scores in the all-class score set, is appropriate for the normal class (i.e., almost none of healthy class samples are rejected), but rejects the significant fraction of known anomaly samples.

To address this issue, we decompose the global score distribution into class-specific distributions using the classifier predictions and assign an individual threshold to each class, as schematically illustrated in Figure \ref{fig:main_scheme}D. 
Thresholds \( \{\tau_{i,p}\}_{i=0}^K \) are defined such that each threshold corresponds to the \((1 - p)\)-quantile of the class-specific anomaly score distribution:
\begin{center}
\(
    \tau_{i,p} = Q_{1 - p}\!\left( s_j \mid \hat{y}_j = c_i \right),
\)
\end{center}
where \( Q_{1 - p}(\cdot) \) denotes the empirical \((1 - p)\)-quantile 
and \( s_j \) is the anomaly score of sample \( x_j \).

\subsection*{Performance on the test set}

We compare several commonly used post-hoc OOD detection methods under different threshold selection strategies. For threshold selection, we fix the proportion of samples predicted as ID (either per class or across all classes, depending on the strategy) to values $p\in[0.95,1.00)$ on the joint training and validation set, using a step size of 0.002. For each combination of method and threshold selection strategy, we select $p$ from the evaluated range which ensures best FPR at $\overline{\mathrm{FNR}}\leqslant 0.25\%$ on the validation set. Despite the ultimate goal being $\overline{\text{FNR}}\approx 0$, our experiments indicate that further reducing $\overline{\text{FNR}}$ leads to excessive number of false positives at the WSI level. Although FPR does not play a key role from a safety-first perspective, it remains important in terms of system usability: a system with too many false alarms is operationally impractical for pathologists. Figures \ref{fig:BER1} and \ref{fig:BER2} in Appendix A1 show \emph{balanced error rate} $\overline{\text{BER}}$ (defined as the mean of $\overline{\text{FNR}}$ and FPR), $\overline{\text{FNR}}$ and FPR across different $p$ on the validation and the test set, respectively. The validation set does not contain OOD samples, therefore $\overline{\text{FNR}}$ is substantially lower on the validation set for all methods. Differences in FPR between the validation and test sets are due to varying glycogen storage in the healthy tissue samples. We observe that our approach (\emph{Adaptive-Maha+}) yields the best $\overline{\text{BER}}$ curve on the test set. We also observe that our approach is reasonably stable with respect to the exact choice of $p$. In the high-percentile regime which is most relevant for safety-critical deployment, $\overline{\text{FNR}}$ in Figure \ref{fig:BER2} changes from e.g. 0.13\% at $p=0.994$ to 0.22\% at $p=0.998$.

Table \ref{tab:ood_results_combined} shows the advantage of Mahalanobis distance-based method (\emph{Maha+}) over competing methods in terms of $\overline{\text{FNR}}$, as well as in terms of misclassified OOD anomalies. The best $\overline{\text{FNR}}$ of 0.16\% is achieved by \emph{Maha+} under the adaptive threshold selection strategy. We stress that the $\overline{\mathrm{FNR}}$ counts only anomalies misclassified as healthy. Under the stricter requirement of correct-class assignment, \emph{Adaptive-Maha+} correctly assigns 93.93\% of ID anomaly and 89.38\% of OOD anomaly pixels to their own class (complementary to the 6.07\% and 10.62\% misclassification rates in Table~\ref{tab:ood_results_combined}, respectively). All other methods misclassify too many OOD samples as healthy and therefore do not meet the safety-first criterion in practice. In particular, KL-Matching, which relies on mean probability vectors as meaningful class prototypes, shows the worst results among all. We hypothesize that, in histopathological data, intra-class heterogeneity and near-OOD overlap cause mean probability vectors to be poor descriptors of the underlying class structure.

Table \ref{tab:ood_results_combined} also compares our adaptive threshold selection strategy with the standard approach in OOD detection, which treats the scores of all ID classes as a single distribution. Both approaches result in comparable $\overline{\text{FNR}}$ and FPR in all methods, with the adaptive strategy achieving a marginal reduction in $\overline{\text{FNR}}$ at the expense of a marginal increase in FPR for some methods. However, for \emph{Adaptive-Maha+} we observe a strong improvement in terms of misclassifying ID anomalies as OOD anomalies, reducing from 28.08\% to 3.64\% (see also confusion matrices in Fig. \ref{fig:matrices} in Appendix A4)
while the percentage of OOD anomalies misclassified as ID anomalies increases from 0.69\% to 10.51\%. In practice, ID anomalies  are likely to occur more often than OOD anomalies, which makes \emph{Adaptive-Maha+} more attractive than \emph{Standard-Maha+} for the purpose of joint segmentation and anomaly detection. In Appendix A1, we additionally provide a comparison of OOD detection methods at an oracle operating point for each method that minimizes $\overline{\text{BER}}$ on the test set (Table \ref{tab:ood_results_combined_2}). We also compare our modified implementation (\emph{Maha+}) with the original implementation of the Mahalanobis distance-based method (\emph{Maha}). At fixed $p$ in Table \ref{tab:ood_results_maha_standard_vs_adaptive}, we observe very similar performance of the two methods, but across different $p$ in Figure \ref{fig:BER2}, \emph{Adaptive-Maha+} shows consistent improvement over \emph{Adaptive-Maha} in terms of $\overline{\text{BER}}$ and FPR at almost identical $\overline{\text{FNR}}$.

In addition to the annotated test set, we tested \emph{Adaptive-Maha+} on 6 full WSIs of liver tissue from the control group, in which no relevant pathological alterations are anticipated. Aside from occasional minor findings -- such as minimal inflammatory cells, mild hepatocellular variation, or technical artifacts -- the tissue is expected to be healthy. The raised false alarms were inspected by a pathologist. We observed that most false alarms were confined to tissue edges or manifested as extremely small regions -- on the order of a fraction of a nucleus -- too small to reflect any genuine morphological change. In practice, these can be largely removed by a lightweight post-processing pipeline consisting of tissue-edge exclusion and a minimum-connected-component-size filter. We would like to emphasize that while achieving $\mathrm{FPR} \approx 0$ is likely infeasible, the false positives generated by our system still impose less workload for a pathologist than a fully manual review. For example, pathologists may perform two modes of work, when assisted by the AI system: 1) a quick confirmatory review of areas predicted healthy, and 2) a detailed inspection of areas predicted as anomalous. Since typically only a small amount of the tissue is anomalous, this would save substantial time. A comprehensive, quantitative pathologist-workload study, including inter-rater agreement and time-to-review measurements on independent WSIs, is a natural next step that we leave to future work.

\begin{table}[H]
\centering
\renewcommand{\arraystretch}{1.3}
\resizebox{\textwidth}{!}{
\begin{tabular}{@{}llccccccc@{}}
\toprule
\addlinespace[0.8em]

\multicolumn{2}{l}{\textbf{Standard threshold selection}}
& MSP\cite{hendrycks2017}
& Max-Logit\cite{hendrycks2022}
& Energy\cite{liu2020energy}
& ReAct\cite{sun2021react}
& KL-M\cite{hendrycks2022}
& OC-SVM\cite{scholkopf2001}
& Maha+ (ours) \\
\cmidrule{3-9}

&
& p = 0.998
& p = 0.998
& p = 0.998
& p = 0.994
& p = 0.998
& p = 0.962
& p = 0.986 \\
\midrule\midrule

\multirow{1}{*}{Anomaly}
& $\overline{\text{FNR}}$
& 3.34\scriptsize$\pm$0.68
& \second{2.43\scriptsize$\pm$0.45}
& \second{2.43\scriptsize$\pm$0.45}
& 3.46\scriptsize$\pm$1.15
& 4.55\scriptsize$\pm$0.84
& 2.60\scriptsize$\pm$0.74
& \best{0.20\scriptsize$\pm$0.06} \\
\midrule

\multirow{3}{*}{Healthy}
& FPR
& 0.32\scriptsize$\pm$0.03
& \second{0.29\scriptsize$\pm$0.02}
& \best{0.28\scriptsize$\pm$0.02}
& 0.30\scriptsize$\pm$0.08
& 0.38\scriptsize$\pm$0.13
& 0.55\scriptsize$\pm$0.13
& 0.32\scriptsize$\pm$0.03 \\
\cmidrule{2-9}

& Misclass.\ as ID-anomaly
& \best{0.25\scriptsize$\pm$0.02}
& \best{0.25\scriptsize$\pm$0.02}
& \best{0.25\scriptsize$\pm$0.02}
& \best{0.25\scriptsize$\pm$0.02}
& \best{0.25\scriptsize$\pm$0.02}
& \best{0.25\scriptsize$\pm$0.02}
& \best{0.25\scriptsize$\pm$0.02} \\

& Misclass.\ as OOD-anomaly
& 0.07\scriptsize$\pm$0.02
& \second{0.04\scriptsize$\pm$0.01}
& \best{0.03\scriptsize$\pm$0.01}
& \second{0.04\scriptsize$\pm$0.06}
& 0.13\scriptsize$\pm$0.11
& 0.30\scriptsize$\pm$0.11
& 0.07\scriptsize$\pm$0.03 \\
\midrule

\multirow{4}{*}{ID-Anomaly}
& Misclassified
& \best{4.56\scriptsize$\pm$0.36}
& 5.17\scriptsize$\pm$0.52
& 5.16\scriptsize$\pm$0.56
& 6.11\scriptsize$\pm$1.81
& \second{4.63\scriptsize$\pm$0.47}
& 36.64\scriptsize$\pm$1.04
& 29.70\scriptsize$\pm$1.40 \\
\cmidrule{2-9}

& Misclass.\ as ID-anomaly
& 2.73\scriptsize$\pm$0.37
& 2.70\scriptsize$\pm$0.37
& 2.70\scriptsize$\pm$0.37
& 2.65\scriptsize$\pm$0.33
& 2.76\scriptsize$\pm$0.37
& \second{1.54\scriptsize$\pm$0.22}
& \best{1.43\scriptsize$\pm$0.20} \\

& Misclass.\ as OOD-anomaly
& \best{1.48\scriptsize$\pm$0.16}
& 2.12\scriptsize$\pm$0.31
& 2.11\scriptsize$\pm$0.31
& 3.11\scriptsize$\pm$1.60
& \second{1.52\scriptsize$\pm$0.37}
& 34.92\scriptsize$\pm$1.03
& 28.08\scriptsize$\pm$1.37 \\

& Misclass.\ as healthy
& 0.35\scriptsize$\pm$0.03
& 0.35\scriptsize$\pm$0.03
& 0.34\scriptsize$\pm$0.03
& 0.34\scriptsize$\pm$0.04
& 0.35\scriptsize$\pm$0.03
& \best{0.18\scriptsize$\pm$0.05}
& \second{0.19\scriptsize$\pm$0.04} \\
\midrule

\multirow{3}{*}{OOD-Anomaly}
& Misclassified
& 69.67\scriptsize$\pm$3.73
& 46.56\scriptsize$\pm$2.01
& \second{46.03\scriptsize$\pm$2.04}
& 66.97\scriptsize$\pm$16.64
& 97.94\scriptsize$\pm$1.49
& 54.10\scriptsize$\pm$5.46
& \best{1.00\scriptsize$\pm$0.95} \\
\cmidrule{2-9}

& Misclass.\ as ID-anomaly
& 39.42\scriptsize$\pm$5.44
& 25.36\scriptsize$\pm$5.23
& \second{24.88\scriptsize$\pm$5.82}
& 35.50\scriptsize$\pm$5.54
& 55.56\scriptsize$\pm$9.82
& 29.70\scriptsize$\pm$3.63
& \best{0.69\scriptsize$\pm$0.56} \\

& Misclass.\ as healthy
& 30.25\scriptsize$\pm$6.72
& 21.20\scriptsize$\pm$4.50
& \second{21.15\scriptsize$\pm$4.51}
& 31.47\scriptsize$\pm$11.48
& 42.38\scriptsize$\pm$8.47
& 24.40\scriptsize$\pm$7.21
& \best{0.30\scriptsize$\pm$0.40} \\

\midrule\midrule
\addlinespace[0.8em]

\multicolumn{2}{l}{\textbf{Adaptive threshold selection (ours)}}
& MSP\cite{hendrycks2017}
& Max-Logit\cite{hendrycks2022}
& Energy\cite{liu2020energy}
& ReAct\cite{sun2021react}
& KL-M\cite{hendrycks2022}
& OC-SVM\cite{scholkopf2001}
& Maha+ (ours) \\
\cmidrule{3-9}

&
& p = 0.998
& p = 0.998
& p = 0.998
& p = 0.996
& p = 0.998
& p = 0.994
& p = 0.996 \\
\midrule\midrule

\multirow{1}{*}{Anomaly}
& $\overline{\text{FNR}}$
& \pairwin{3.32\scriptsize$\pm$0.67}
& \pairwin{2.39\scriptsize$\pm$0.42}
& \pairwin{\second{2.38\scriptsize$\pm$0.42}}
& \pairwin{3.44\scriptsize$\pm$1.16}
& \pairwin{4.40\scriptsize$\pm$0.79}
& 2.68\scriptsize$\pm$0.83
& \pairwin{\best{0.16\scriptsize$\pm$0.04}} \\
\midrule

\multirow{3}{*}{Healthy}
& FPR
& 0.33\scriptsize$\pm$0.03
& \second{0.29\scriptsize$\pm$0.02}
& \best{0.28\scriptsize$\pm$0.02}
& 0.31\scriptsize$\pm$0.11
& 0.45\scriptsize$\pm$0.19
& 0.57\scriptsize$\pm$0.30
& 0.35\scriptsize$\pm$0.03 \\
\cmidrule{2-9}

& Misclass.\ as ID-anomaly
& \best{0.25\scriptsize$\pm$0.02}
& \best{0.25\scriptsize$\pm$0.02}
& \best{0.25\scriptsize$\pm$0.02}
& \best{0.25\scriptsize$\pm$0.02}
& \best{0.25\scriptsize$\pm$0.02}
& \best{0.25\scriptsize$\pm$0.02}
& \best{0.25\scriptsize$\pm$0.02} \\

& Misclass.\ as OOD-anomaly
& 0.07\scriptsize$\pm$0.02
& \second{0.04\scriptsize$\pm$0.01}
& \best{0.03\scriptsize$\pm$0.01}
& 0.06\scriptsize$\pm$0.08
& 0.20\scriptsize$\pm$0.17
& 0.32\scriptsize$\pm$0.29
& 0.09\scriptsize$\pm$0.02 \\
\midrule

\multirow{4}{*}{ID-Anomaly}
& Misclassified
& \second{5.04\scriptsize$\pm$0.57}
& 5.85\scriptsize$\pm$0.81
& 5.71\scriptsize$\pm$0.80
& \pairwin{5.30\scriptsize$\pm$0.84}
& \pairwin{\best{4.04\scriptsize$\pm$0.49}}
& \pairwin{17.10\scriptsize$\pm$2.84}
& \pairwin{6.07\scriptsize$\pm$0.85} \\
\cmidrule{2-9}

& Misclass.\ as ID-anomaly
& \pairwin{2.72\scriptsize$\pm$0.36}
& \pairwin{2.68\scriptsize$\pm$0.35}
& 2.70\scriptsize$\pm$0.36
& \pairwin{2.62\scriptsize$\pm$0.37}
& 2.79\scriptsize$\pm$0.39
& \second{2.32\scriptsize$\pm$0.32}
& \best{2.26\scriptsize$\pm$0.22} \\

& Misclass.\ as OOD-anomaly
& \second{1.97\scriptsize$\pm$0.22}
& 2.82\scriptsize$\pm$0.54
& 2.67\scriptsize$\pm$0.53
& \pairwin{2.35\scriptsize$\pm$0.63}
& \pairwin{\best{0.92\scriptsize$\pm$0.15}}
& \pairwin{14.50\scriptsize$\pm$2.76}
& \pairwin{3.64\scriptsize$\pm$0.79} \\

& Misclass.\ as healthy
& 0.35\scriptsize$\pm$0.03
& 0.35\scriptsize$\pm$0.03
& 0.34\scriptsize$\pm$0.03
& 0.34\scriptsize$\pm$0.04
& \pairwin{0.33\scriptsize$\pm$0.03}
& \second{0.28\scriptsize$\pm$0.04}
& \pairwin{\best{0.17\scriptsize$\pm$0.04}} \\
\midrule

\multirow{3}{*}{OOD-Anomaly}
& Misclassified
& \pairwin{67.53\scriptsize$\pm$3.00}
& \pairwin{40.71\scriptsize$\pm$4.13}
& \pairwin{\second{40.19\scriptsize$\pm$3.66}}
& \pairwin{61.22\scriptsize$\pm$13.23}
& \pairwin{97.31\scriptsize$\pm$1.82}
& 56.31\scriptsize$\pm$8.27
& \best{10.62\scriptsize$\pm$3.60} \\
\cmidrule{2-9}

& Misclass.\ as ID-anomaly
& \pairwin{37.48\scriptsize$\pm$4.37}
& \pairwin{19.90\scriptsize$\pm$4.11}
& \pairwin{\second{19.50\scriptsize$\pm$3.92}}
& \pairwin{29.89\scriptsize$\pm$3.93}
& 56.33\scriptsize$\pm$9.22
& 32.03\scriptsize$\pm$5.15
& \best{10.51\scriptsize$\pm$3.57} \\

& Misclass.\ as healthy
& \pairwin{30.05\scriptsize$\pm$6.62}
& \pairwin{20.80\scriptsize$\pm$4.17}
& \pairwin{\second{20.69\scriptsize$\pm$4.19}}
& \pairwin{31.34\scriptsize$\pm$11.60}
& \pairwin{40.98\scriptsize$\pm$7.96}
& \pairwin{24.27\scriptsize$\pm$8.26}
& \pairwin{\best{0.11\scriptsize$\pm$0.11}} \\

\bottomrule
\end{tabular}
}
\caption{Comparison of OOD detection methods under two thresholding strategies: \textbf{Standard} (upper) and \textbf{Adaptive} (lower). Each combination of method and threshold selection strategy is reported at $p$ which ensures best FPR at $\overline{\mathrm{FNR}}\leqslant 0.25\%$ on the validation set. All reported metrics are expressed as percentages, whereas $p$ values are unitless fractions. The table groups performance by the ground truth category. Under \emph{Anomaly}, $\overline{\mathrm{FNR}}$ is the safety-critical rate of anomalous pixels predicted as healthy. Under \emph{Healthy}, FPR is the overall false alarm rate, split into false alarms flagged as ID anomaly and false alarms flagged as OOD anomaly. Under \emph{ID-Anomaly}, \emph{Misclassified} is the fraction of ID anomaly pixels not assigned to their correct class, split into: a different known class, OOD class, or healthy; the last row (\emph{Misclass.\ as healthy}) is again safety-critical. Under \emph{OOD-Anomaly}, \emph{Misclassified} is the fraction of OOD pixels not flagged as OOD, split into misclassifying as ID anomaly and missing as healthy (safety-critical). Values in the Adaptive section highlighted in gray \pairwin{} indicate improvement over the corresponding Standard result. Best values are \best{bold}; second-best are \second{underlined}.}
\label{tab:ood_results_combined}
\end{table}

\subsection*{Detection of OOD anomalies on the whole-slide image}

\begin{figure}[H]
\centering
\includegraphics[width=1\linewidth]{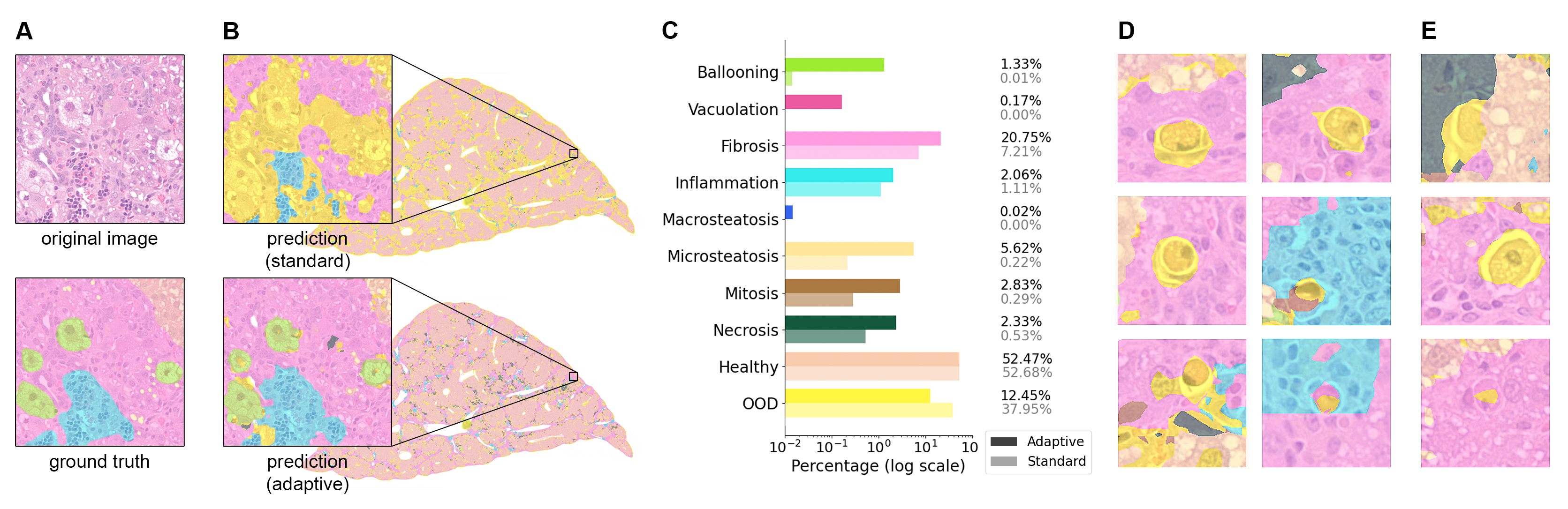}
\caption{Whole-slide image analysis of mouse liver stained in H\&E from a CCl\textsubscript{4}-induced disease model to study metabolic dysfunction-associated steatohepatitis (MASH). \textbf{A}. Example areas of WSI  showing the original image and the ground truth annotated by a pathologist. \textbf{B}. The same areas with overlaid predictions using a single threshold (\emph{Standard-Maha+}) and the adaptive thresholds per class (\emph{Adaptive-Maha+}). \textbf{C}. Proportion of liver tissue area occupied by each anomaly type in comparison for the single threshold (\emph{Standard-Maha+}) and the adaptive thresholds (\emph{Adaptive-Maha+}). \textbf{D}. Apoptotic cells detected as OOD (yellow)  from our annotated test set with the \emph{Adaptive-Maha+} approach. \textbf{E}. Apoptotic cells detected as OOD (yellow) not present in the test set and confirmed by a certified pathologist (\emph{Adaptive-Maha+} approach).}
\label{fig:WSI_apopt}
\end{figure}

\begin{figure}[H]
\centering
\includegraphics[width=1\linewidth]{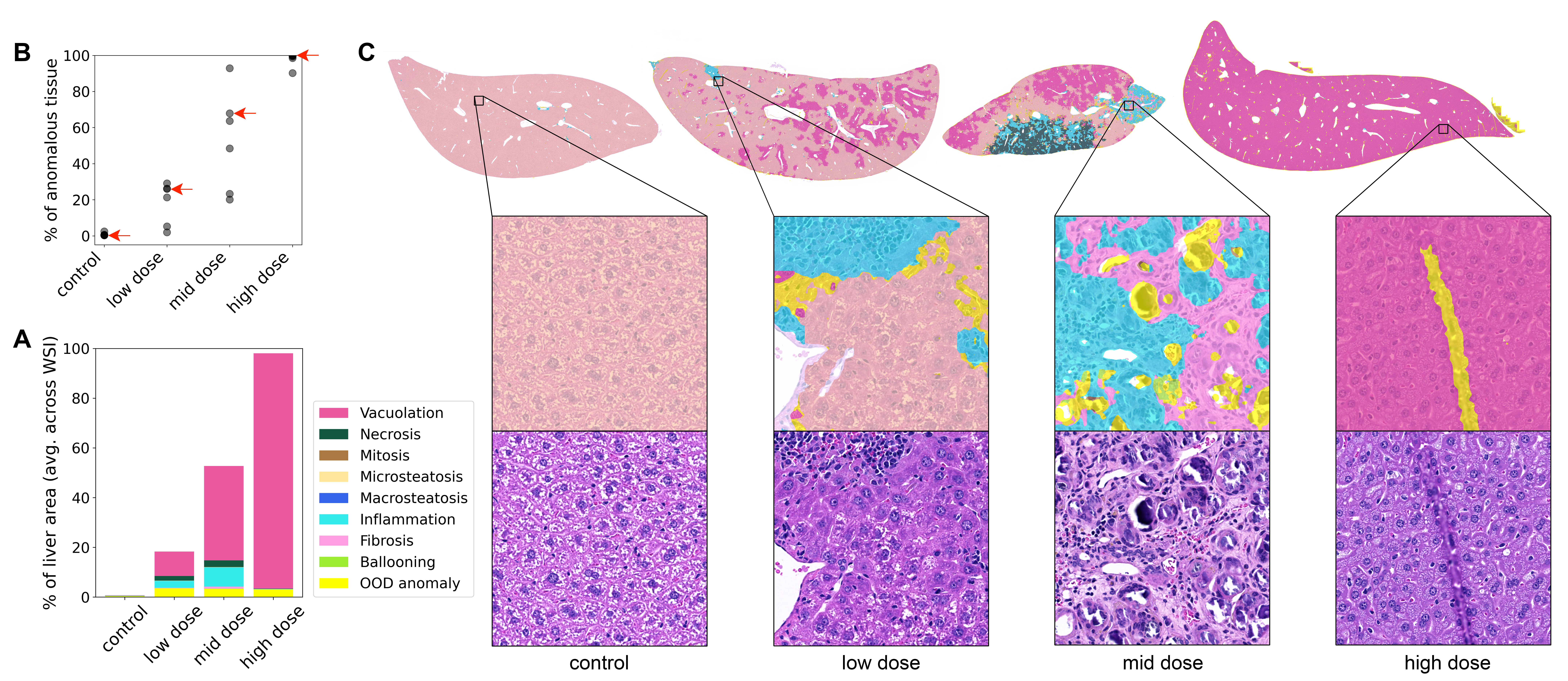}
\caption{Detection of adverse drug reactions in a preclinical toxicological liver study. Our approach (\emph{Adaptive-Maha+}) disentangles dose-dependent toxicologic effects from vehicle-related and incidental anomalies at pixel level. \textbf{A}. Percentage of liver area occupied by different anomalies (averaged across WSIs) is shown for each dose group. Results correspond to pathological alterations and were confirmed by a pathologist. \textbf{B}. Each dot corresponds to a single WSI and shows total percentage of anomalies in the WSI. Four arrows correspond to four WSI examples given in C. \textbf{C}. Examples of detected anomalies in different groups. Cytoplasmic vacuolation shows dose dependency, while other anomaly types do not.}
\label{fig:tox_study}
\end{figure}

Figure \ref{fig:WSI_apopt} illustrates an example of an H\&E-stained liver WSI from an induced damage model, processed by the proposed system at fixed $p=0.996$. Images on the left (Fig. \ref{fig:WSI_apopt}A and \ref{fig:WSI_apopt}B) show example areas of original image and overlays indicating the different types of predicted anomalies. We observe that for sufficiently rare and spatially localized known anomalies like ballooning, the adaptive threshold selection yields better segmentation, while the standard threshold selection tends to reject such classes as OOD. Figure \ref{fig:WSI_apopt}C shows the percentage of liver area occupied by each anomaly type, supporting the observation that adaptive threshold selection yields less unnecessary OOD predictions.

Existing labeled data for Apoptosis were not used during training in order to simulate the OOD anomaly class. Apoptosis represents a particularly challenging case, as it manifests at the single-cell level: apoptotic cells are frequently smaller than typical hepatocytes, making them difficult to identify both algorithmically and visually. Despite this challenge, the system successfully detected apoptotic cells (Fig. \ref{fig:WSI_apopt}D). Notably, several previously unlabeled regions were also flagged as OOD by the model and subsequently confirmed as apoptotic cells by an expert pathologist (Fig. \ref{fig:WSI_apopt}E).

We also observed false predictions, with some necrotic and mitotic regions segmented incorrectly. False necrosis detections were primarily located in healthy areas exhibiting reduced glycogen content. This can be attributed to the denser and darker cytoplasm compared to the predominantly glycogen-rich healthy hepatocytes represented in training dataset. False mitosis detections often occurred when nuclei were sectioned tangentially near the apical plane, resulting in a smaller and hyperchromatic appearance resembling mitotic figures.

\subsection*{Validation of the method on a preclinical toxicological liver study}

We validated our approach using material from a preclinical toxicology study, where H\&E-stained liver tissues were previously diagnosed by an expert pathologist. In this study, mice received increasing doses of a compound being developed to treat a disease unrelated to liver pathology. Histological review by a toxicopathologist showed a compound-induced, dose-dependent increase of hepatocellular cytoplasmic vacuolation. In addition, lesions  without clear dose dependency, such as large amounts of hepatocellular degeneration, necrosis, inflammation, mineralization, as well as pigment deposition were diagnosed.

The results of the analysis from our system at fixed $p=0.996$ are shown in Figure \ref{fig:tox_study}A. Results are analogous to a pathologist’s visual assessment, while additionally supplying detailed quantitative measurements. A clear dose-dependent increase is observed for cytoplasmic vacuolation, whereas the remaining anomalies do not display this trend. Figure \ref{fig:tox_study}B provides an aggregated overview (total percentage of anomalous tissue) for separate WSIs. In Figure \ref{fig:tox_study}C, concrete examples from the control, low-, mid- and high-dose groups (left to right) are shown; the concrete examples are denoted by arrows in Figure \ref{fig:tox_study}B. In the low-dose group, healthy tissue is shown, detected partly as healthy and partly as OOD. This is not an error, but rather a reflection of the fact that some healthy hepatocytes are glycogen-depleted and therefore differ from most of the training data. In the mid-dose group, a complex anomaly is observed, consisting of known components (fibrotic tissue and inflammatory cells) alongside unseen pathological variations classified as OOD. In the high-dose group, the main dose-dependent pathological finding -- cytoplasmic vacuolation -- is present, along with a scanning artifact labeled as OOD, as it belongs to our unknown anomaly dataset.

The same toxicology study was previously analyzed using a binary AD framework by Zingman et al.\cite{zingman2024learning} that distinguished only between healthy and anomalous tissue. While this previous approach successfully identified overall pathological burden in a comparable manner, it did not provide insight into underlying pathologies, and the spatial resolution was limited (i.e., no segmentation). This demonstrates the practical advantages of our system compared to the previous approach.

\section*{Conclusion}

We developed an anomaly detection framework for  whole-slide images to support histopathological screening in preclinical studies, a step toward faster and more reliable decision-making in drug development. The system leverages realistically available data -- healthy tissue and known pathology examples -- while remaining capable of identifying our held-out pathology categories as out-of-distribution anomalies. In accordance with the safety-first principle, the framework achieves an extremely low false negative rate, ensuring that virtually no pathological regions are misclassified as healthy. This is essential for toxicology workflows, where missed findings may allow unsafe compounds to progress into human trials.

Our AI-driven framework accurately detects subtle hepatocellular changes and localizes classic dose-dependent toxicologic lesions including cytoplasmic vacuolation and necrosis, while also flagging atypical tissue regions such as glycogen-depleted hepatocytes and pathologies from our held-out OOD dataset: apoptotic cells and  staining/processing artifacts. Validation on rodent liver WSIs with known toxicologic findings confirmed robust detection performance across multiple anomaly classes.

We note that the present study is limited to mouse liver tissue, and therefore direct generalization to other organs or species should be made with caution. Differences in tissue morphology, staining characteristics, and species-specific pathology may require dataset-specific adaptation and retraining to maintain performance.

To further enhance generalizability and translational value, future work should incorporate diverse tissues, multiple species, scanning platforms, and continuous pathologist feedback, alongside continual-learning strategies designed to maintain specificity while prioritizing further reduction of false negative rates. Additionally, more held-out categories should be added in order to make OOD class representative of the full morphological diversity of rare lesions that a deployed system may encounter. Integration with latest vision foundation models\cite{simeoni2025dinov3}, optimally adapted to histopathology, may further enhance detection performance. Collectively, these advances will strengthen AI-assisted histopathology as a consistent and comprehensive tool for early toxicity assessment, supporting earlier and more reliable decision-making in preclinical studies.

\section*{Methods}

\subsection*{Microscopy}

We build our dataset from WSIs which were acquired using a Zeiss AxioScan scanner (Carl Zeiss, Jena, Germany) equipped with a 20x objective, yielding a native resolution of 0.221 $\mu$m per pixel. All WSIs were acquired from mouse liver tissue sections stained with hematoxylin and eosin (H\&E) following standard histological procedures. Subsequently, the WSIs were downsampled by a factor of 2, resulting in a final spatial resolution of 0.442 $\mu$m per pixel.

\subsection*{Animals}

No new animal experiments were done. Only historic image data from liver sections from previous animal studies was reanalyzed. In these previous experiments, Mice (C57BL/6JRj) at different ages were used. Animals included healthy controls, a CCl\textsubscript{4} MASH model, and a previous toxicologic study. Animals were maintained in accordance with German national guidelines, legal regulations and the guidelines of the Association for Accreditation of Laboratory Animal Care. Experiments were performed after permission from the Regierungspräsidium Tübingen, Germany.

\subsection*{Data annotation}

Per-pixel annotations were performed for all classes using the Halo Link platform (Indica Labs, USA). Manual annotation of whole-slide images was carried out with polygonal tools to delineate lesion boundaries. Each class of anomaly, as well as healthy tissue regions, was carefully reviewed and annotated by an experienced veterinary pathologist, in accordance with established toxicologic pathology criteria and the INHAND (International Harmonization of Nomenclature and Diagnostic Criteria for Lesions in Rats and Mice) guidelines\cite{thoolen2010proliferative}. The annotation process was iterative, involving cross-verification and quality checks to ensure accuracy and consistency across the dataset. Due to the large size and complexity of WSIs, we adopted a partial annotation strategy to ensure practical feasibility, i.e., the dataset contains pixelwise annotations only for selected regions of interest within each whole-slide image.

\subsection*{Classes of annotated tissue}

The dataset includes \textit{healthy} tissue class and nine classes of common lesions: \textit{apoptosis}, \textit{ballooning}, \textit{cytoplasmic vacuolation}, \textit{fibrosis}, \textit{inflammation}, \textit{macrosteatosis}, \textit{microsteatosis}, \textit{mitosis}, and \textit{necrosis}. Additionally, an \textit{artifact} class is included to account for regions affected by tissue processing or staining imperfections. Furthermore, we also included samples representing the vascular system (i.e., cross-sections of blood vessels) and background regions (e.g., at the tissue edge or at sites of tissue disruption) under the label \textit{no tissue}. Although these classes do not represent pathological lesions, they provide contextual information that can assist pathologists in tissue assessment. Fig. \ref{fig:tSNE} in Appendix A5 shows a t-SNE visualization of pixel representations of the test data.

\subsection*{Data splits}

Our dataset comprises 742 WSIs obtained from 44 preclinical studies. To prevent data leakage and better approximate real-world evaluation conditions, the test set was constructed from studies excluded from the training or validation sets. The training and validation sets originate from the same studies but contain liver sections from different animals to ensure enough independence between training and validation set. In order to simulate OOD anomaly detection, we allocated two anomaly classes -- apoptosis and artifact -- to our joint OOD class, and therefore use the corresponding samples exclusively at the test time. This choice was motivated by the scarcity of annotated samples for these classes, rendering effective feature learning infeasible. Moreover, the apoptosis class is particularly challenging, as it represents lesions at the single-cell level, making it a valuable case for evaluating the sensitivity of a segmentation-based approach. Apoptosis also represents a near-OOD case, because it shares certain morphological and feature-space similarities with other anomalies, such as single-cell necrosis or mitosis. In contrast, artifact samples can be considered far-OOD. Morphologically, they do not resemble any biological anomaly class: they may appear as dark contamination spots, severely blurred regions, or folded tissue where the underlying cellular structure is entirely disrupted. These characteristics potentially place artifacts far from biologically meaningful classes in the feature space. Anomaly scores provided by our method (\emph{Maha+}) further corroborate our assumptions: the mean test-set score is lowest (i.e., farthest from the predicted class mean) for the artifact class (-7246), followed by the apoptosis class (-4387), and is substantially higher for ID classes (-1000).

The annotated polygons differ in both size and quantity depending on the anomaly class. The OOD class in the test set contains 78 polygons with a mean area of 14,032 pixels (2,741 $\mu$m$^2$) per polygon. For the ID classes, we have on average 1,101 polygons per class with a mean area of 324,832 pixels (63,461 $\mu$m$^2$) per polygon. However, inter-class variation in these metrics is high, with the number of polygons per class ranging from 327 to 3,470, and the mean polygon area ranging from 790 to 2,416,819 pixels (154 to 472,161 $\mu$m$^2$). Table \ref{tab:dataset_stats} in Appendix A2 shows polygon counts and total pixel counts for the training, validation, and test sets. Inter-class variation in the total number of pixels spans up to 3.5 orders of magnitude. Therefore, we apply a stratified dataset split based on the Wasserstein distance. This approach ensures that the distribution of pixels across classes in the training, validation, and test sets remains as similar as possible. Specifically, we compute the Wasserstein distance between class-wise pixel count histograms and iteratively adjust the split to achieve the smallest overall discrepancy. Nevertheless, a perfectly balanced split cannot be achieved due to our restriction of not mixing studies or animals between subsets, as well as the inherently uneven distribution of lesion types across studies. We divide the dataset with a target ratio of 70\%, 15\%, and 15\% for the training, validation, and test sets, respectively; however, the achieved proportions are approximate, reflecting the aforementioned limitations.

\subsection*{Tiling strategy}

Our ViT backbone converts an image into a sequence of patch embeddings with a native patch size of 14×14 px, so all spatial resolutions in our pipeline are chosen as integer multiples of this patch size to facilitate straightforward interpolation of the model outputs back to the original image resolution. Accordingly, we train and evaluate our method using 672×672 px tiles extracted from WSIs, whose central 252×252 px regions are arranged on a non-overlapping grid and the surrounding context windows overlap.

\subsection*{Fine-tuning and inference}

The DINOv2 \cite{oquab2023dinov2} encoder with ViT-Base/14 and embedding dimension $d=768$ was initialized with pretrained weights provided by Facebook Research. All original weights of the DINOv2 backbone (including patch embeddings, transformer blocks, and normalization layers) were kept frozen, while only the task-specific components were fine-tuned: LoRA\cite{hu2022lora} modules with rank $r=3$ were applied to the attention layers, and the final linear segmentation head was updated. LoRA parameter ablations are reported in Table \ref{tab:lora_rank_ablation} in Appendix 1. The training objective was the categorical cross-entropy loss weighted by inverse class frequencies to mitigate strong class imbalance in our data. Pixels corresponding to unlabeled regions were ignored during optimization. We fine-tuned the model for 50 epochs using the AdamW optimizer with a learning rate of $3\times10^{-4}$ and batch size 12. The learning rate was reduced by a factor of 0.5 when the validation loss did not improve for two consecutive epochs. The best-performing model was selected based on the mean Intersection-over-Union (IoU) 
computed on a held-out validation set. During training, non-empty 252×252 px crops were sampled randomly from 672×672 px tiles at each epoch. We define a crop as non-empty if: for the microsteatosis class, it contains at least one annotated pixel; for the vacuolation and mitosis classes, at least 0.5\% of its area is occupied by the annotated pixels; for the rest of the classes, at least 1\% of its area is occupied by the annotated pixels. 

During the OOD detector calibration procedure and inference, we used a stride of 84 px for spatial shift averaging within the central 252×252 px region, thus averaging across 36 shifts. Table \ref{tab:inference_benchmark} in Appendix A1 compares shift-averaged inference with single-pass inference and reports the average time per tile as well as the expected time per WSI. The empirical overhead of 26.4× is lower than the theoretical overhead of 36×, as the per-tile inference time includes a fixed cost (image loading, preprocessing, and GPU transfer) that does not scale with the number of shifts. Although the WSI inference time increases substantially for 36 shifts, from 3.2 to 83 minutes, Table \ref{tab:ood_results_combined_noshift} in Appendix 1 confirms that the advantage of \emph{Adaptive-Maha+} over other methods is preserved in single-pass mode. In resource-constrained deployment settings, we envision a two-stage workflow in which a fast single-pass inference is performed first. WSIs with numerous raised alarms, which already provide strong evidence of toxicity, are then reviewed based on these coarse predictions alone, whereas WSIs with absent or subtle findings undergo the shift-averaging procedure to obtain more precise and robust predictions.

Thresholds, i.e., class-specific \((1 - p)\)-score quantiles, were estimated on the joint training and validation set for values of $p$ in the range $[0.95,1.00)$ with a step size of 0.002. The final operating value of $p$ was selected based on validation-set performance as the value yielding the lowest FPR while satisfying $\overline{\mathrm{FNR}}\leqslant 0.25\%$.

We implemented the experiments in the PyTorch deep learning framework \cite{paszke2019pytorch}. We trained 5 models with different random seeds to randomize the tile cropping and the order of feeding tiles to the network. For the test set, we provide mean and standard error values over 5 trained models, which allows for more reliable performance comparisons.

\newpage

\bibliography{sample}

\section*{Funding}

Olga Graf is supported by the AI \& Data Science Fellowship Post Doc program from Boehringer Ingelheim with the University of Tübingen, Germany.

\section*{Author contributions statement}

O.G. did the machine learning approach development, coding, data analysis, and provided the first draft of the manuscript. D.P. provided annotation of histopathological slides and histopathological expertise. P.G. provided study data, and provided technical input. C.L. provided histopathological expertise. M.H. supervised O.G. about the machine learning method development in his research group. F.H. initiated and led the project from Boehringer Ingelheim. All authors reviewed the manuscript. 

\section*{Competing interests}

The authors declare no competing interests.

\section*{Data availability}

The dataset generated during the current study is available in the Open Science Framework repository, \url{https://osf.io/mkvct/}. Source code of the method in the paper is available at GitHub, \url{https://github.com/Boehringer-Ingelheim/anomaly-segmentation-and-detection}.

\newpage

\section*{Appendix}

\subsection*{A1. Ablation studies}

\begin{figure}[H]
\centering
\includegraphics[width=1\linewidth]{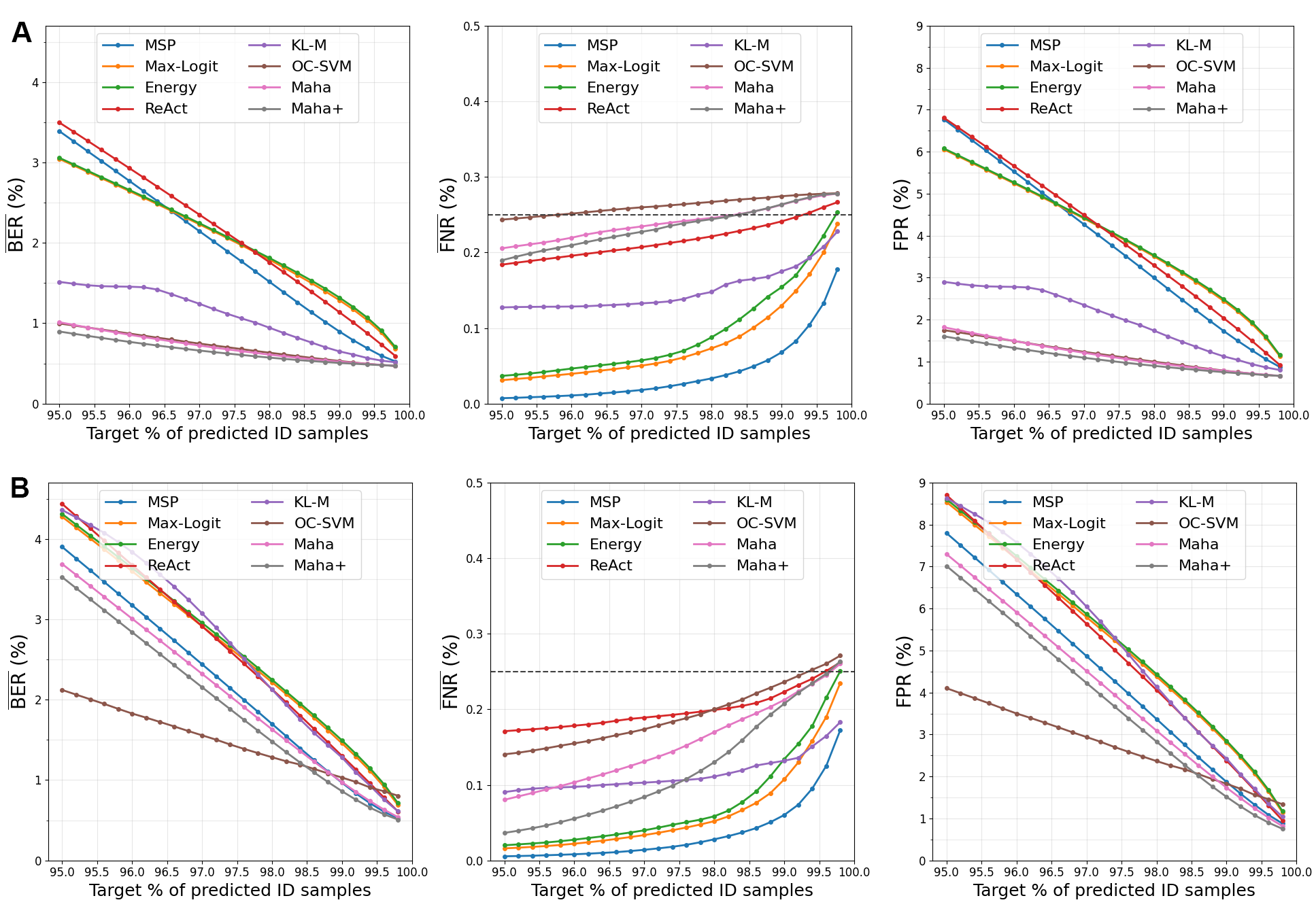}
\caption{Performance of OOD detection methods on the validation set under \textbf{Standard} (\textbf{A}) and \textbf{Adaptive} (\textbf{B}) threshold selection 
across various target percentages of predicted ID samples. Balanced error rate ($\overline{\text{BER}}$), false negative rate ($\overline{\text{FNR}}$) and false positive rate (FPR) are evaluated on the validation set at different target percentages determined on the training+validation set. Validation set contains no OOD samples. The horizontal dashed line indicates the threshold at 0.25\%. The plots show mean values over 5 trained models.}
\label{fig:BER1}
\end{figure}

\newpage

\begin{figure}[H]
\centering
\includegraphics[width=1\linewidth]{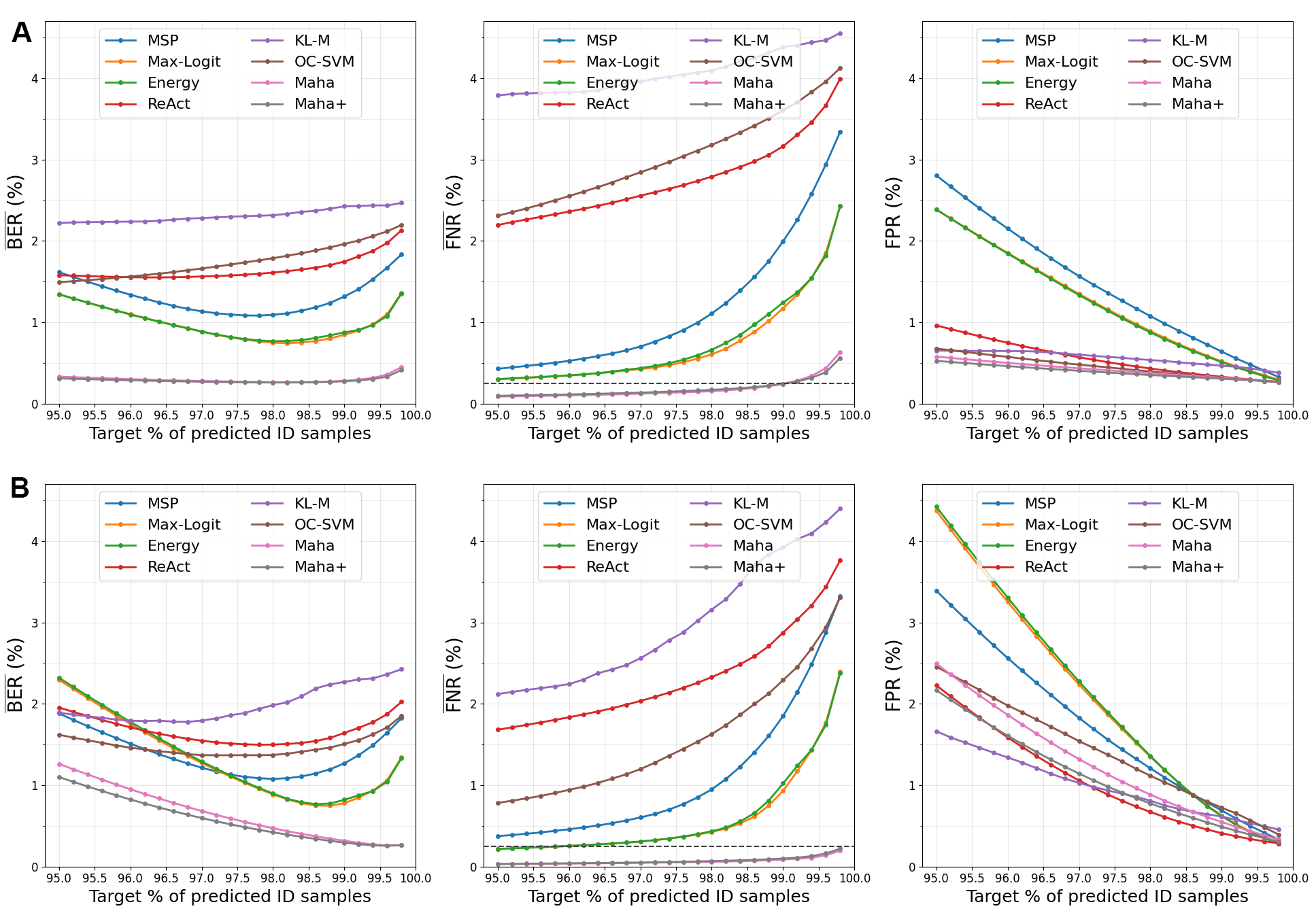}
\caption{Performance of OOD detection methods on the test set under \textbf{Standard} (\textbf{A}) and \textbf{Adaptive} (\textbf{B}) threshold selection 
across various target percentages of predicted ID samples. Balanced error rate ($\overline{\text{BER}}$), false negative rate ($\overline{\text{FNR}}$) and false positive rate (FPR) are evaluated on the test set at different target percentages determined on the training+validation set. The horizontal dashed line indicates the threshold at 0.25\%. The plots show mean values over 5 trained models.}
\label{fig:BER2}
\end{figure}

\begin{table}[H]
\centering
\renewcommand{\arraystretch}{1.3}
\resizebox{0.74\textwidth}{!}{
\begin{tabular}{@{}ll p{0.8cm} cccc@{}}
\toprule
\cmidrule{1-7}
& & & $r=1$ & $r=3$ & $r=8$ & $r=16$ \\
\midrule\midrule

\multirow{1}{*}{Anomaly}
& $\overline{\text{FNR}}$
&
& 0.25\scriptsize$\pm$0.11
& \underline{0.16\scriptsize$\pm$0.04}
& 0.17\scriptsize$\pm$0.10
& \textbf{0.13\scriptsize$\pm$0.05} \\

\midrule
\multirow{3}{*}{Healthy}
& FPR
&
& \textbf{0.32\scriptsize$\pm$0.03}
& 0.35\scriptsize$\pm$0.03
& \underline{0.34\scriptsize$\pm$0.09}
& 0.43\scriptsize$\pm$0.21 \\

\cmidrule{2-7}
& Misclass.\ as ID-anomaly
&
& \textbf{0.21\scriptsize$\pm$0.03}
& 0.25\scriptsize$\pm$0.02
& 0.24\scriptsize$\pm$0.06
& \underline{0.23\scriptsize$\pm$0.05} \\

& Misclass.\ as OOD-anomaly
&
& \underline{0.10\scriptsize$\pm$0.02}
& \textbf{0.09\scriptsize$\pm$0.02}
& 0.11\scriptsize$\pm$0.03
& 0.21\scriptsize$\pm$0.20 \\

\midrule
\multirow{4}{*}{ID-Anomaly}
& Misclassified
&
& \textbf{4.45\scriptsize$\pm$0.48}
& \underline{6.07\scriptsize$\pm$0.85}
& 6.49\scriptsize$\pm$1.20
& 6.72\scriptsize$\pm$0.54 \\

\cmidrule{2-7}
& Misclass.\ as (wrong) ID anomaly
&
& 2.10\scriptsize$\pm$0.30
& 2.26\scriptsize$\pm$0.22
& \textbf{2.00\scriptsize$\pm$0.31}
& \underline{2.06\scriptsize$\pm$0.37} \\

& Misclass.\ as OOD-anomaly
&
& \textbf{2.08\scriptsize$\pm$0.26}
& \underline{3.64\scriptsize$\pm$0.79}
& 4.31\scriptsize$\pm$1.42
& 4.53\scriptsize$\pm$0.91 \\

& Misclass.\ as healthy
&
& 0.27\scriptsize$\pm$0.11
& \underline{0.17\scriptsize$\pm$0.04}
& 0.18\scriptsize$\pm$0.11
& \textbf{0.13\scriptsize$\pm$0.06} \\

\midrule
\multirow{3}{*}{OOD-Anomaly}
& Misclassified
&
& 8.13\scriptsize$\pm$1.07
& 10.62\scriptsize$\pm$3.60
& \textbf{6.71\scriptsize$\pm$0.86}
& \underline{8.07\scriptsize$\pm$3.14} \\

\cmidrule{2-7}
& Misclass.\ as ID-anomaly
&
& 8.07\scriptsize$\pm$1.04
& 10.51\scriptsize$\pm$3.57
& \textbf{6.65\scriptsize$\pm$0.79}
& \underline{7.99\scriptsize$\pm$3.16} \\

& Misclass.\ as healthy
&
& \underline{0.06\scriptsize$\pm$0.05}
& 0.11\scriptsize$\pm$0.11
& \textbf{0.05\scriptsize$\pm$0.08}
& 0.07\scriptsize$\pm$0.10 \\

\bottomrule
\end{tabular}
}
\caption{
LoRA-rank ablation for the proposed \emph{Adaptive-Maha+} detector at $p=0.996$. All reported metrics are expressed as percentages. Best values are shown in \textbf{bold}, and second-best values are \underline{underlined}. The table shows values over 5 trained models.}
\label{tab:lora_rank_ablation}
\end{table}

\begin{table}[H]
\centering
\renewcommand{\arraystretch}{1.3}
\resizebox{\textwidth}{!}{
\begin{tabular}{@{}llccccccc@{}}
\toprule
\addlinespace[0.8em]

\multicolumn{2}{l}{\textbf{Standard threshold selection}}
& MSP\cite{hendrycks2017} 
& Max-Logit\cite{hendrycks2022} 
& Energy\cite{liu2020energy} 
& ReAct\cite{sun2021react} 
& KL-M\cite{hendrycks2022} 
& OC-SVM\cite{scholkopf2001} 
& Maha+ \\
\cmidrule{3-9}

&
& p = 0.998
& p = 0.998
& p = 0.998
& p = 0.994
& p = 0.998
& p = 0.962
& p = 0.986 \\
\midrule\midrule

\multirow{1}{*}{Anomaly}
& $\overline{\text{FNR}}$
& 3.27\scriptsize$\pm$0.89
& 2.90\scriptsize$\pm$0.63
& 2.90\scriptsize$\pm$0.59
& 3.32\scriptsize$\pm$1.06
& 3.90\scriptsize$\pm$0.81
& \second{1.64\scriptsize$\pm$0.31}
& \best{0.65\scriptsize$\pm$0.28} \\
\midrule

\multirow{3}{*}{Healthy}
& FPR
& 0.37\scriptsize$\pm$0.06
& \best{0.34\scriptsize$\pm$0.04}
& \best{0.34\scriptsize$\pm$0.04}
& \second{0.36\scriptsize$\pm$0.08}
& 0.42\scriptsize$\pm$0.08
& 0.49\scriptsize$\pm$0.09
& \best{0.34\scriptsize$\pm$0.04} \\
\cmidrule{2-9}

& Misclass.\ as ID-anomaly
& \best{0.28\scriptsize$\pm$0.03}
& \second{0.31\scriptsize$\pm$0.04}
& \second{0.31\scriptsize$\pm$0.04}
& \second{0.31\scriptsize$\pm$0.04}
& 0.32\scriptsize$\pm$0.04
& 0.32\scriptsize$\pm$0.05
& \second{0.31\scriptsize$\pm$0.05} \\

& Misclass.\ as OOD-anomaly
& 0.09\scriptsize$\pm$0.03
& \best{0.03\scriptsize$\pm$0.01}
& \best{0.03\scriptsize$\pm$0.00}
& \second{0.05\scriptsize$\pm$0.04}
& 0.10\scriptsize$\pm$0.07
& 0.18\scriptsize$\pm$0.05
& \best{0.03\scriptsize$\pm$0.02} \\
\midrule

\multirow{4}{*}{ID-Anomaly}
& Misclassified
& \best{4.76\scriptsize$\pm$0.45}
& \second{5.05\scriptsize$\pm$0.46}
& 5.08\scriptsize$\pm$0.48
& 6.07\scriptsize$\pm$1.20
& 5.47\scriptsize$\pm$0.86
& 31.23\scriptsize$\pm$0.43
& 19.42\scriptsize$\pm$1.68 \\
\cmidrule{2-9}

& Misclass.\ as ID-anomaly
& 3.17\scriptsize$\pm$0.36
& 3.03\scriptsize$\pm$0.35
& 3.06\scriptsize$\pm$0.36
& 3.01\scriptsize$\pm$0.35
& 3.64\scriptsize$\pm$0.42
& \second{2.31\scriptsize$\pm$0.32}
& \best{1.83\scriptsize$\pm$0.26} \\

& Misclass.\ as OOD-anomaly
& \best{1.15\scriptsize$\pm$0.13}
& 1.58\scriptsize$\pm$0.17
& 1.56\scriptsize$\pm$0.19
& 2.60\scriptsize$\pm$1.29
& \second{1.33\scriptsize$\pm$0.71}
& 28.56\scriptsize$\pm$0.63
& 17.15\scriptsize$\pm$1.52 \\

& Misclass.\ as healthy
& \second{0.44\scriptsize$\pm$0.11}
& \second{0.44\scriptsize$\pm$0.12}
& 0.46\scriptsize$\pm$0.12
& 0.46\scriptsize$\pm$0.12
& 0.49\scriptsize$\pm$0.05
& \best{0.36\scriptsize$\pm$0.13}
& \second{0.44\scriptsize$\pm$0.12} \\
\midrule

\multirow{3}{*}{OOD-Anomaly}
& Misclassified
& 76.95\scriptsize$\pm$2.84
& 62.09\scriptsize$\pm$3.93
& 61.66\scriptsize$\pm$4.40
& 77.39\scriptsize$\pm$11.02
& 97.24\scriptsize$\pm$2.46
& \second{40.06\scriptsize$\pm$3.52}
& \best{5.75\scriptsize$\pm$3.75} \\
\cmidrule{2-9}

& Misclass.\ as ID-anomaly
& 48.21\scriptsize$\pm$6.68
& 37.03\scriptsize$\pm$8.83
& 36.75\scriptsize$\pm$9.17
& 48.36\scriptsize$\pm$2.15
& 62.70\scriptsize$\pm$8.87
& \second{26.96\scriptsize$\pm$3.18}
& \best{3.25\scriptsize$\pm$1.24} \\

& Misclass.\ as healthy
& 28.75\scriptsize$\pm$8.64
& 25.06\scriptsize$\pm$5.99
& 24.91\scriptsize$\pm$5.61
& 29.03\scriptsize$\pm$10.12
& 34.54\scriptsize$\pm$8.04
& \second{13.10\scriptsize$\pm$2.80}
& \best{2.51\scriptsize$\pm$2.60} \\

\midrule\midrule
\addlinespace[0.8em]

\multicolumn{2}{l}{\textbf{Adaptive threshold selection (ours)}}
& MSP\cite{hendrycks2017} 
& Max-Logit\cite{hendrycks2022} 
& Energy\cite{liu2020energy} 
& ReAct\cite{sun2021react} 
& KL-M\cite{hendrycks2022} 
& OC-SVM\cite{scholkopf2001} 
& Maha+ \\
\cmidrule{3-9}

&
& p = 0.998
& p = 0.998
& p = 0.998
& p = 0.996
& p = 0.998
& p = 0.994
& p = 0.996 \\
\midrule\midrule

\multirow{1}{*}{Anomaly}
& $\overline{\text{FNR}}$
& \pairwin{2.93\scriptsize$\pm$0.89}
& \pairwin{2.45\scriptsize$\pm$0.54}
& \pairwin{2.46\scriptsize$\pm$0.50}
& \pairwin{3.18\scriptsize$\pm$1.10}
& \pairwin{3.83\scriptsize$\pm$0.78}
& \second{2.06\scriptsize$\pm$0.46}
& \pairwin{\best{0.38\scriptsize$\pm$0.12}} \\
\midrule

\multirow{3}{*}{Healthy}
& FPR
& 0.42\scriptsize$\pm$0.07
& \second{0.37\scriptsize$\pm$0.05}
& \best{0.36\scriptsize$\pm$0.04}
& 0.38\scriptsize$\pm$0.12
& 0.57\scriptsize$\pm$0.15
& \pairwin{0.45\scriptsize$\pm$0.13}
& \best{0.36\scriptsize$\pm$0.04} \\
\cmidrule{2-9}

& Misclass.\ as ID-anomaly
& \best{0.30\scriptsize$\pm$0.04}
& \second{0.31\scriptsize$\pm$0.04}
& 0.32\scriptsize$\pm$0.04
& \second{0.31\scriptsize$\pm$0.04}
& 0.32\scriptsize$\pm$0.04
& 0.32\scriptsize$\pm$0.04
& 0.32\scriptsize$\pm$0.04 \\

& Misclass.\ as OOD-anomaly
& 0.12\scriptsize$\pm$0.04
& \second{0.05\scriptsize$\pm$0.01}
& \best{0.04\scriptsize$\pm$0.01}
& 0.07\scriptsize$\pm$0.08
& 0.25\scriptsize$\pm$0.13
& \pairwin{0.12\scriptsize$\pm$0.09}
& \best{0.04\scriptsize$\pm$0.01} \\
\midrule

\multirow{4}{*}{ID-Anomaly}
& Misclassified
& \pairwin{\best{4.50\scriptsize$\pm$0.42}}
& \pairwin{\second{4.55\scriptsize$\pm$0.40}}
& \pairwin{4.57\scriptsize$\pm$0.40}
& \pairwin{4.85\scriptsize$\pm$0.43}
& \pairwin{4.88\scriptsize$\pm$0.58}
& \pairwin{15.89\scriptsize$\pm$1.36}
& \pairwin{5.24\scriptsize$\pm$0.45} \\
\cmidrule{2-9}

& Misclass.\ as ID-anomaly
& 3.30\scriptsize$\pm$0.37
& \second{3.26\scriptsize$\pm$0.35}
& 3.31\scriptsize$\pm$0.35
& 3.37\scriptsize$\pm$0.45
& 3.68\scriptsize$\pm$0.43
& \best{3.10\scriptsize$\pm$0.37}
& 3.40\scriptsize$\pm$0.42 \\

& Misclass.\ as OOD-anomaly
& \pairwin{\second{0.82\scriptsize$\pm$0.07}}
& \pairwin{0.90\scriptsize$\pm$0.08}
& \pairwin{0.85\scriptsize$\pm$0.08}
& \pairwin{1.05\scriptsize$\pm$0.19}
& \pairwin{\best{0.73\scriptsize$\pm$0.28}}
& \pairwin{12.33\scriptsize$\pm$1.27}
& \pairwin{1.46\scriptsize$\pm$0.22} \\

& Misclass.\ as healthy
& \pairwin{\best{0.38\scriptsize$\pm$0.09}}
& \pairwin{\second{0.39\scriptsize$\pm$0.12}}
& \pairwin{0.41\scriptsize$\pm$0.12}
& \pairwin{0.43\scriptsize$\pm$0.14}
& \pairwin{0.46\scriptsize$\pm$0.05}
& 0.46\scriptsize$\pm$0.13
& \pairwin{\best{0.38\scriptsize$\pm$0.13}} \\
\midrule

\multirow{3}{*}{OOD-Anomaly}
& Misclassified
& 78.36\scriptsize$\pm$2.52
& \pairwin{61.53\scriptsize$\pm$4.36}
& \pairwin{61.57\scriptsize$\pm$5.04}
& \pairwin{76.06\scriptsize$\pm$10.23}
& 97.87\scriptsize$\pm$2.40
& \second{49.39\scriptsize$\pm$3.98}
& \best{21.07\scriptsize$\pm$5.12} \\
\cmidrule{2-9}

& Misclass.\ as ID-anomaly
& 52.45\scriptsize$\pm$7.21
& 40.55\scriptsize$\pm$6.04
& 40.62\scriptsize$\pm$6.36
& \pairwin{48.18\scriptsize$\pm$3.84}
& 63.71\scriptsize$\pm$9.36
& \second{32.89\scriptsize$\pm$3.67}
& \best{20.62\scriptsize$\pm$4.99} \\

& Misclass.\ as healthy
& \pairwin{25.91\scriptsize$\pm$8.55}
& \pairwin{20.98\scriptsize$\pm$4.92}
& \pairwin{20.95\scriptsize$\pm$4.49}
& \pairwin{27.88\scriptsize$\pm$10.29}
& \pairwin{34.16\scriptsize$\pm$7.65}
& \second{16.50\scriptsize$\pm$3.99}
& \pairwin{\best{0.46\scriptsize$\pm$0.31}} \\

\bottomrule
\end{tabular}
}
\caption{Performance of single-pass inference (i.e., without averaging across shifts) across OOD detection methods under two thresholding strategies: \textbf{Standard} (upper) and \textbf{Adaptive} (lower). Each combination of method and threshold selection strategy is reported at same $p$ as in Table 1 in the main text. All reported metrics are expressed as percentages, whereas $p$ values are unitless fractions. Values in the Adaptive section highlighted in gray \pairwin{} indicate improvement (lower mean) over the corresponding Standard result. Best values are \best{bold}; second-best are \second{underlined}.} \label{tab:ood_results_combined_noshift} \end{table}

\begin{table}[H]
\centering
\renewcommand{\arraystretch}{1.3}
\resizebox{\textwidth}{!}{
\begin{tabular}{@{}llccccccc@{}}
\toprule
\addlinespace[0.8em]

\multicolumn{2}{l}{\textbf{Standard threshold selection}} 
 & MSP\cite{hendrycks2017} 
 & Max-Logit\cite{hendrycks2022} 
 & Energy\cite{liu2020energy} 
 & ReAct\cite{sun2021react} 
 & KL-M\cite{hendrycks2022} 
 & OC-SVM\cite{scholkopf2001} 
 & Maha+ (ours) \\
\cmidrule{3-9}
 & 
 & p = 0.978 & p = 0.982 & p = 0.980 & p = 0.964 & p = 0.950 & p = 0.950 & p = 0.980 \\
\midrule\midrule

\multirow{1}{*}{Anomaly}
& $\overline{\text{FNR}}$ 
& 0.99\scriptsize$\pm$0.30
& 0.68\scriptsize$\pm$0.20
& \second{0.66\scriptsize$\pm$0.20}
& 2.43\scriptsize$\pm$1.01
& 3.79\scriptsize$\pm$0.91
& 2.31\scriptsize$\pm$0.67
& \best{0.17\scriptsize$\pm$0.04} \\
\midrule

\multirow{3}{*}{Healthy}
& FPR 
& 1.17\scriptsize$\pm$0.22
& 0.81\scriptsize$\pm$0.09
& 0.87\scriptsize$\pm$0.10
& 0.67\scriptsize$\pm$0.35
& \second{0.65\scriptsize$\pm$0.28}
& 0.68\scriptsize$\pm$0.16
& \best{0.35\scriptsize$\pm$0.04} \\
\cmidrule{2-9}
& Misclass.\ as ID-anomaly 
& \best{0.25\scriptsize$\pm$0.02} 
& \best{0.25\scriptsize$\pm$0.02} 
& \best{0.25\scriptsize$\pm$0.02} 
& \best{0.25\scriptsize$\pm$0.02} 
& \best{0.25\scriptsize$\pm$0.01} 
& \best{0.25\scriptsize$\pm$0.02}
& \best{0.25\scriptsize$\pm$0.02} \\
& Misclass.\ as OOD-anomaly 
& 0.92\scriptsize$\pm$0.21
& 0.56\scriptsize$\pm$0.09
& 0.62\scriptsize$\pm$0.10
& 0.42\scriptsize$\pm$0.34
& \second{0.40\scriptsize$\pm$0.27}
& 0.43\scriptsize$\pm$0.14
& \best{0.10\scriptsize$\pm$0.04} \\
\midrule

\multirow{4}{*}{ID-Anomaly}
& Misclassified  
& \best{13.42\scriptsize$\pm$1.02}
& \second{13.63\scriptsize$\pm$1.44}
& 14.49\scriptsize$\pm$1.52
& 16.35\scriptsize$\pm$5.22
& 26.01\scriptsize$\pm$1.18
& 43.36\scriptsize$\pm$0.94
& 35.30\scriptsize$\pm$1.52 \\
\cmidrule{2-9}
& Misclass.\ as ID-anomaly  
& 2.25\scriptsize$\pm$0.22
& 2.21\scriptsize$\pm$0.16
& 2.17\scriptsize$\pm$0.15
& 2.23\scriptsize$\pm$0.27
& 2.07\scriptsize$\pm$0.24
& \second{1.34\scriptsize$\pm$0.20}
& \best{1.27\scriptsize$\pm$0.19} \\
& Misclass.\ as OOD-anomaly 
& \best{10.89\scriptsize$\pm$0.97}
& \second{11.14\scriptsize$\pm$1.34}
& 12.04\scriptsize$\pm$1.44
& 13.83\scriptsize$\pm$5.27
& 23.69\scriptsize$\pm$1.20
& 41.86\scriptsize$\pm$0.92
& 33.86\scriptsize$\pm$1.42 \\
& Misclass.\ as healthy 
& 0.29\scriptsize$\pm$0.02
& 0.29\scriptsize$\pm$0.03
& 0.28\scriptsize$\pm$0.03
& 0.29\scriptsize$\pm$0.05
& 0.25\scriptsize$\pm$0.04
& \best{0.16\scriptsize$\pm$0.04}
& \second{0.17\scriptsize$\pm$0.03} \\
\midrule

\multirow{3}{*}{OOD-Anomaly}
& Misclassified  
& 17.03\scriptsize$\pm$2.59
& 9.49\scriptsize$\pm$2.44
& \second{9.30\scriptsize$\pm$2.85}
& 42.96\scriptsize$\pm$16.35
& 80.05\scriptsize$\pm$4.37
& 48.02\scriptsize$\pm$5.34
& \best{0.46\scriptsize$\pm$0.41} \\
\cmidrule{2-9}
& Misclass.\ as ID-anomaly 
& 9.68\scriptsize$\pm$0.88
& \second{5.31\scriptsize$\pm$1.61}
& 5.24\scriptsize$\pm$2.19
& 21.31\scriptsize$\pm$6.59
& 44.45\scriptsize$\pm$8.85
& 26.34\scriptsize$\pm$3.65
& \best{0.32\scriptsize$\pm$0.23} \\
& Misclass.\ as healthy 
& 7.34\scriptsize$\pm$3.03
& 4.18\scriptsize$\pm$1.91
& \second{4.06\scriptsize$\pm$1.94}
& 21.65\scriptsize$\pm$10.03
& 35.60\scriptsize$\pm$9.16
& 21.68\scriptsize$\pm$6.56
& \best{0.14\scriptsize$\pm$0.20} \\

\midrule\midrule
\addlinespace[0.8em]

\multicolumn{2}{l}{\textbf{Adaptive threshold selection (ours)}} 
 & MSP\cite{hendrycks2017} 
 & Max-Logit\cite{hendrycks2022} 
 & Energy\cite{liu2020energy} 
 & ReAct\cite{sun2021react} 
 & KL-M\cite{hendrycks2022} 
 & OC-SVM\cite{scholkopf2001} 
 & Maha+ (ours) \\
\cmidrule{3-9}
 & 
 & p = 0.980 & p = 0.988 & p = 0.986 & p = 0.978 & p = 0.968 & p = 0.972 & p = 0.996 \\
\midrule\midrule

\multirow{1}{*}{Anomaly}
& $\overline{\text{FNR}}$ 
& \pairwin{0.95\scriptsize$\pm$0.29}
& 0.75\scriptsize$\pm$0.21
& \second{0.66\scriptsize$\pm$0.20}
& \pairwin{2.26\scriptsize$\pm$1.15}
& \pairwin{2.48\scriptsize$\pm$0.79}
& \pairwin{1.27\scriptsize$\pm$0.46}
& \pairwin{\best{0.16\scriptsize$\pm$0.04}} \\
\midrule

\multirow{3}{*}{Healthy}
& FPR 
& 1.21\scriptsize$\pm$0.23
& \pairwin{0.75\scriptsize$\pm$0.09}
& 0.88\scriptsize$\pm$0.12
& \second{0.74\scriptsize$\pm$0.44}
& 1.08\scriptsize$\pm$0.34
& 1.46\scriptsize$\pm$0.69
& \best{0.35\scriptsize$\pm$0.03} \\
\cmidrule{2-9}
& Misclass.\ as ID-anomaly 
& \second{0.25\scriptsize$\pm$0.02} 
& \second{0.25\scriptsize$\pm$0.02} 
& \second{0.25\scriptsize$\pm$0.02} 
& \second{0.25\scriptsize$\pm$0.02} 
& \second{0.25\scriptsize$\pm$0.02} 
& \pairwin{\best{0.24\scriptsize$\pm$0.01}}
& \second{0.25\scriptsize$\pm$0.02} \\
& Misclass.\ as OOD-anomaly 
& 0.95\scriptsize$\pm$0.22
& \pairwin{\second{0.49\scriptsize$\pm$0.09}}
& 0.62\scriptsize$\pm$0.11
& \second{0.49\scriptsize$\pm$0.42}
& 0.83\scriptsize$\pm$0.33
& 1.22\scriptsize$\pm$0.69
& \pairwin{\best{0.09\scriptsize$\pm$0.02}} \\
\midrule

\multirow{4}{*}{ID-Anomaly}
& Misclassified  
& \pairwin{12.23\scriptsize$\pm$0.83}
& \pairwin{10.65\scriptsize$\pm$1.22}
& \pairwin{11.29\scriptsize$\pm$1.29}
& \pairwin{10.33\scriptsize$\pm$1.53}
& \pairwin{\second{8.48\scriptsize$\pm$0.61}}
& \pairwin{31.68\scriptsize$\pm$2.07}
& \pairwin{\best{6.07\scriptsize$\pm$0.85}} \\
\cmidrule{2-9}
& Misclass.\ as ID-anomaly  
& 2.37\scriptsize$\pm$0.25
& 2.42\scriptsize$\pm$0.24
& 2.39\scriptsize$\pm$0.24
& 2.32\scriptsize$\pm$0.38
& 2.63\scriptsize$\pm$0.34
& \best{1.76\scriptsize$\pm$0.31}
& \second{2.26\scriptsize$\pm$0.22} \\
& Misclass.\ as OOD-anomaly 
& \pairwin{9.57\scriptsize$\pm$0.63}
& \pairwin{7.94\scriptsize$\pm$1.05}
& \pairwin{8.61\scriptsize$\pm$1.12}
& \pairwin{7.73\scriptsize$\pm$1.43}
& \pairwin{\second{5.71\scriptsize$\pm$0.62}}
& \pairwin{29.72\scriptsize$\pm$2.09}
& \pairwin{\best{3.64\scriptsize$\pm$0.79}} \\
& Misclass.\ as healthy 
& 0.29\scriptsize$\pm$0.02
& 0.29\scriptsize$\pm$0.03
& 0.28\scriptsize$\pm$0.03
& \pairwin{0.28\scriptsize$\pm$0.07}
& \pairwin{\best{0.15\scriptsize$\pm$0.03}}
& 0.20\scriptsize$\pm$0.03
& \second{0.17\scriptsize$\pm$0.04} \\
\midrule

\multirow{3}{*}{OOD-Anomaly}
& Misclassified  
& 19.51\scriptsize$\pm$2.83
& 11.23\scriptsize$\pm$2.80
& \best{9.86\scriptsize$\pm$2.94}
& \pairwin{37.00\scriptsize$\pm$12.49}
& \pairwin{74.05\scriptsize$\pm$6.98}
& \pairwin{28.67\scriptsize$\pm$4.26}
& \second{10.62\scriptsize$\pm$3.60} \\
\cmidrule{2-9}
& Misclass.\ as ID-anomaly 
& 12.61\scriptsize$\pm$0.79
& \second{6.35\scriptsize$\pm$1.23}
& \best{5.83\scriptsize$\pm$1.54}
& \pairwin{16.94\scriptsize$\pm$3.16}
& 50.62\scriptsize$\pm$9.85
& \pairwin{17.74\scriptsize$\pm$3.00}
& 10.51\scriptsize$\pm$3.57 \\
& Misclass.\ as healthy 
& \pairwin{6.90\scriptsize$\pm$2.89}
& 4.88\scriptsize$\pm$2.04
& \pairwin{\second{4.03\scriptsize$\pm$1.87}}
& \pairwin{20.07\scriptsize$\pm$11.23}
& \pairwin{23.43\scriptsize$\pm$7.92}
& \pairwin{10.92\scriptsize$\pm$4.51}
& \pairwin{\best{0.11\scriptsize$\pm$0.11}} \\
\bottomrule
\end{tabular}
}
\caption{Comparison of OOD detection methods under two thresholding strategies: \textbf{Standard} (upper) and \textbf{Adaptive} (lower). Each combination of method and threshold selection strategy is reported at its best achievable performance in terms of $\overline{\text{BER}}$ and the corresponding $p$ is given. All reported metrics are expressed as percentages, whereas $p$ values are unitless fractions. Values in the Adaptive section highlighted in gray \pairwin{} indicate improvement (lower mean) over the corresponding Standard result. Best values are \best{bold}; second-best are \second{underlined}.}
\label{tab:ood_results_combined_2}
\end{table}

\begin{table}[H]
\centering
\renewcommand{\arraystretch}{1.3}
\resizebox{0.95\textwidth}{!}{
\begin{tabular}{@{}ll p{1cm} cc p{1cm} cc@{}}
\toprule
\addlinespace[0.8em]
 & 
 & 
 & \multicolumn{2}{c}{\textbf{Standard threshold selection}}
 & 
 & \multicolumn{2}{c}{\textbf{Adaptive threshold selection (ours)}} \\
\cmidrule{1-8}
 & 
 & 
 & Maha\cite{lee2018}
 & Maha+ (ours)
 & 
 & Maha\cite{lee2018}
 & Maha+ (ours) \\
\cmidrule{4-8}
 & 
 & 
 & p = 0.982
 & p = 0.980
 & 
 & p = 0.996
 & p = 0.996 \\
\midrule\midrule

\multirow{1}{*}{Anomaly}
& $\overline{\text{FNR}}$
& 
& \best{0.16\scriptsize$\pm$0.07}
& 0.17\scriptsize$\pm$0.04
& 
& \pairwin{\best{0.14\scriptsize$\pm$0.04}}
& \pairwin{0.16\scriptsize$\pm$0.04} \\

\midrule
\multirow{3}{*}{Healthy}
& FPR
& 
& 0.36\scriptsize$\pm$0.05
& \best{0.35\scriptsize$\pm$0.04}
& 
& 0.38\scriptsize$\pm$0.05
& \best{0.35\scriptsize$\pm$0.03} \\

\cmidrule{2-8}
& Misclass.\ as ID-anomaly
& 
& \best{0.25\scriptsize$\pm$0.02}
& \best{0.25\scriptsize$\pm$0.02}
& 
& \best{0.25\scriptsize$\pm$0.02}
& \best{0.25\scriptsize$\pm$0.02} \\

& Misclass.\ as OOD-anomaly
& 
& 0.11\scriptsize$\pm$0.04
& \best{0.10\scriptsize$\pm$0.04}
& 
& 0.13\scriptsize$\pm$0.04
& \pairwin{\best{0.09\scriptsize$\pm$0.02}} \\

\midrule
\multirow{4}{*}{ID-Anomaly}
& Misclassified
& 
& 38.41\scriptsize$\pm$2.03
& \best{35.30\scriptsize$\pm$1.52}
& 
& \pairwin{\best{5.90\scriptsize$\pm$0.68}}
& \pairwin{6.07\scriptsize$\pm$0.85} \\

\cmidrule{2-8}
& Misclass.\ as ID-anomaly
& 
& \best{1.22\scriptsize$\pm$0.21}
& 1.27\scriptsize$\pm$0.19
& 
& 2.27\scriptsize$\pm$0.24
& \best{2.26\scriptsize$\pm$0.22} \\

& Misclass.\ as OOD-anomaly
& 
& 37.03\scriptsize$\pm$1.96
& \best{33.86\scriptsize$\pm$1.42}
& 
& \pairwin{\best{3.49\scriptsize$\pm$0.60}}
& \pairwin{3.64\scriptsize$\pm$0.79} \\

& Misclass.\ as healthy
& 
& \best{0.15\scriptsize$\pm$0.05}
& 0.17\scriptsize$\pm$0.03
& 
& \pairwin{\best{0.14\scriptsize$\pm$0.04}}
& 0.17\scriptsize$\pm$0.04 \\

\midrule
\multirow{3}{*}{OOD-Anomaly}
& Misclassified
& 
& 0.59\scriptsize$\pm$0.49
& \best{0.46\scriptsize$\pm$0.41}
& 
& 11.18\scriptsize$\pm$2.99
& \best{10.62\scriptsize$\pm$3.60} \\

\cmidrule{2-8}
& Misclass.\ as ID-anomaly
& 
& 0.33\scriptsize$\pm$0.18
& \best{0.32\scriptsize$\pm$0.23}
& 
& 11.05\scriptsize$\pm$2.97
& \best{10.51\scriptsize$\pm$3.57} \\

& Misclass.\ as healthy
& 
& 0.26\scriptsize$\pm$0.33
& \best{0.14\scriptsize$\pm$0.20}
& 
& \pairwin{0.12\scriptsize$\pm$0.13}
& \pairwin{\best{0.11\scriptsize$\pm$0.11}} \\

\bottomrule
\end{tabular}
}
\caption{
Comparison of original (\emph{Maha}) and modified (\emph{Maha+}) Mahalanobis distance-based methods under \textbf{Standard} and \textbf{Adaptive} threshold selection. Each combination of method and threshold selection strategy is reported at its best achievable performance in terms of $\overline{\text{BER}}$ and the corresponding $p$ is given. All reported metrics are expressed as percentages, whereas $p$ values are unitless fractions. Values highlighted in gray \pairwin{} indicate improvement (lower mean) over the corresponding Standard result.
Best values are \best{bold}.
}
\label{tab:ood_results_maha_standard_vs_adaptive}
\end{table}

\begin{table}[h]
\centering
\small
\begin{tabular}{lcccc}
\hline
\textbf{Mode} & \textbf{Mean (s)} & \textbf{Std (s)} & \textbf{Overhead} & \textbf{Est. time/WSI (min)}\\
\hline
Single-pass & 0.073 & 0.011 & 1.0$\times$ & $\approx 3.2$ \\
Shift-averaged (36 passes)                & 1.913 & 0.076 & 26.4$\times$ &  $\approx 83$ \\
\hline
\end{tabular}

\caption{Inference time comparison between single-pass and shift-averaged modes, averaged over 50 tiles. Estimated WSI inference times are computed using the median number of tiles per slide. Computations were performed on a NVIDIA A100-PCIE-40GB GPU.}
\label{tab:inference_benchmark}
\end{table}

\subsection*{A2. Dataset size}

\begin{table*}[h]
\centering
\begin{tabular}{l|rr|rr|rr}
\hline
\textbf{Class} & \multicolumn{2}{c|}{\textbf{Training Set}} & \multicolumn{2}{c|}{\textbf{Validation Set}} & \multicolumn{2}{c}{\textbf{Test Set}} \\
 & Polygons & Pixels (M) & Polygons & Pixels (M) & Polygons & Pixels (M) \\
\hline
apoptosis      & -   & -  & -  & -  & 29  & 0.16 \\
artifact       & -   & -  & -  & -  & 49  & 0.90 \\
ballooning     & 326  & 8.52  & 63  & 1.73  & 86  & 2.84 \\
vacuolation    & 296  & 39.00 & 147 & 6.46  & 98  & 11.20 \\
fibrosis       & 203  & 35.77 & 55  & 14.07 & 69  & 4.26 \\
inflammation   & 381  & 35.23 & 107 & 7.53  & 149 & 6.96 \\
macrosteatosis & 1010 & 60.38 & 258 & 12.25 & 555 & 18.09 \\
microsteatosis & 1662 & 1.55  & 492 & 0.44  & 1316& 0.75 \\
mitosis        & 678  & 12.26 & 214 & 3.34  & 167 & 1.80 \\
necrosis       & 473  & 157.52& 142 & 40.66 & 239 & 15.19 \\
no tissue      & 751  & 104.85& 124 & 13.94 & 160 & 23.22 \\
normal         & 469  & 1289.94 & 164 & 273.89 & 138 & 298.83 \\
\hline
\end{tabular}
\caption{Polygon counts and total pixel counts (in millions) per class for train, validation, and test sets.}
\label{tab:dataset_stats}
\end{table*}

\subsection*{A3. Averaging across spatial shifts}\label{sec:avg_shifts}

Assume \( I \in \mathbb{R}^{T \times T \times 3} \) is an extended image tile and 
\( P \) are the corresponding per-pixel class probability maps, where each \( P(x) \in \Delta^K \) lies on the \((K+1)\)-dimensional probability simplex.

Image tile \( I \) is evaluated using $t\times t$ ($t<T$) sliding window over a set of \emph{spatial shifts}:
\begin{center}
\(
\mathcal{S} = \{\, (x,y) \mid x,y \in [0, T - t], \; \text{ where } x,y 
\text{ are sampled every } k \text{ pixels}\,\}.
\)
\end{center}

Namely, for each spatial shift $(x,y)$, 
a sub-tile
\(
I_{x,y} = \{\, I(i,j) \mid i \in [y,y+t),\; j \in [x,x+t) \,\}
\)
is fed into the model, yielding corresponding probabilities \( P_{x,y}\).

Now consider the \emph{central region} of the extended tile,
\(\Omega \subset [0,t]\times[0,t]\) -- the area that will be used for aggregating results. We restrict $T<3t$ such that each shift contributes to $\Omega$. The aggregated prediction for pixel $u\in\Omega$ is computed as the mean over all shifted predictions covering that pixel:
\begin{center}
\(
\bar{P}(u) = \frac{1}{|\mathcal{S}|} 
\sum_{(x,y) \in \mathcal{S}} P_{x,y}(u).
\)
\end{center}
The final predicted label is then obtained as $\hat{y}(u) = \arg\max_{c_i} \bar{P}_{c_i}(u).$

Each sub-tile $I_{x,y}$ yields feature embeddings, 
\(H_{x,y} \in \mathbb{R}^{d \times t' \times t'}\), where $d$ is the embedding dimension and $t'\times t'$ denotes the spatial resolution of the latent feature map. Feature embeddings are aggregated in the same manner, yielding averaged features:
\begin{center}
\(
\bar{H}(u') = \frac{1}{|\mathcal{S}|} 
\sum_{(x,y) \in \mathcal{S}} H_{x,y}(u'),
\)
\end{center}
where $u'\in\Omega'$ belongs to the central region in the latent space. 

In per-pixel anomaly score calculation, based on such feature representations, anomaly score maps can be subsequently interpolated to match the size of prediction maps.

\subsection*{A4. Confusion matrices}

\begin{figure}[H]
\centering
\includegraphics[width=1\linewidth]{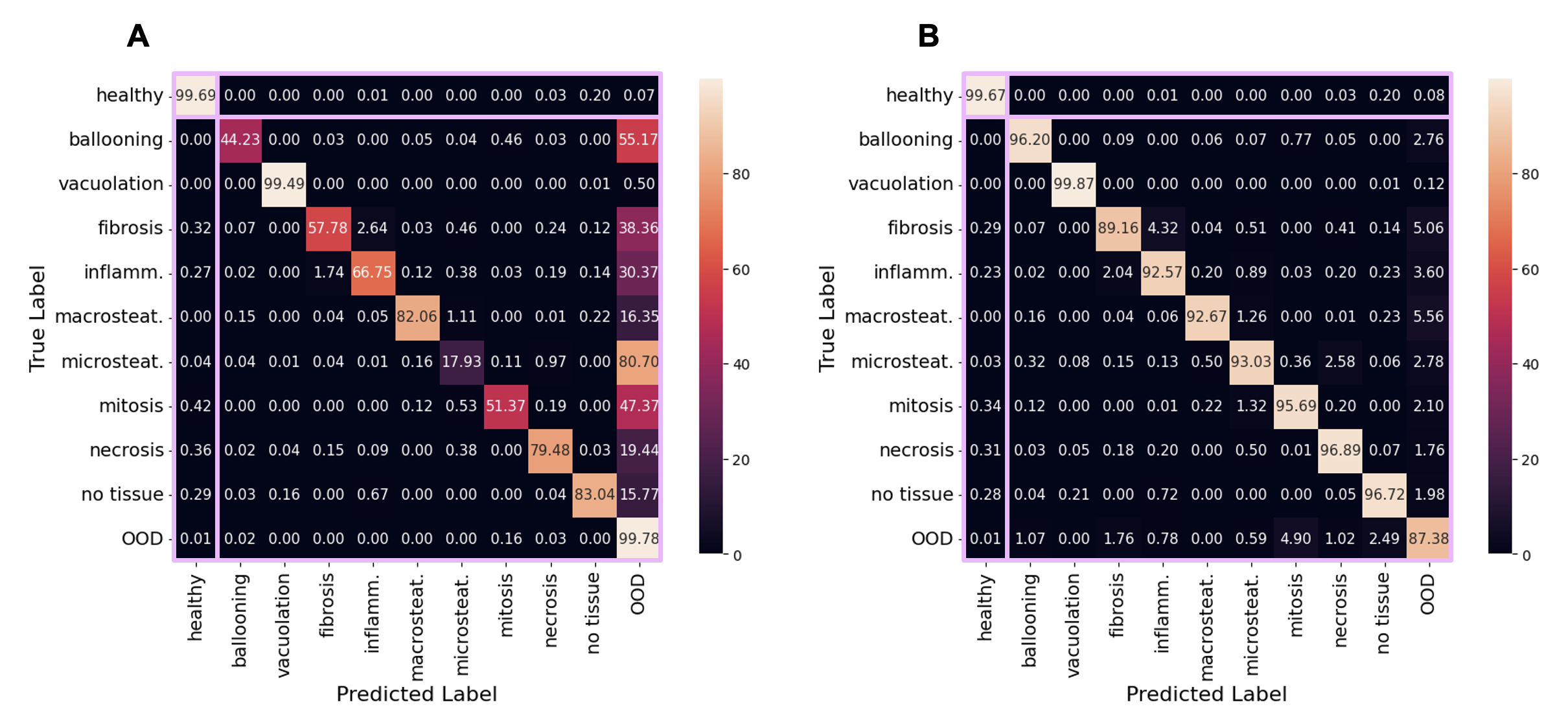}
\caption{Extended confusion matrices for unified segmentation and anomaly detection, evaluated with \emph{Standard-Maha+} ($p=0.980$) and \emph{Adaptive-Maha+} ($p=0.996$) on joint training and validation set; the choice of $p$ reflects the best achievable performance in terms of $\overline{\text{BER}}$. \textbf{A}. \textbf{Standard} threshold selection. A substantial fraction of ID anomalies is misclassified as OOD, especially in rare or spatially highly localized anomaly classes like microsteatosis, ballooning, mitosis. \textbf{B}. \textbf{Adaptive} threshold selection. ID anomalies are classified correctly with high precision, OOD anomalies are classified correctly with less precision than in A, but are nonetheless not missed.}
\label{fig:matrices}
\end{figure}

\subsection*{A5. t-SNE of pixel features}
\begin{figure}[H]
\centering
\includegraphics[width=0.65\linewidth]{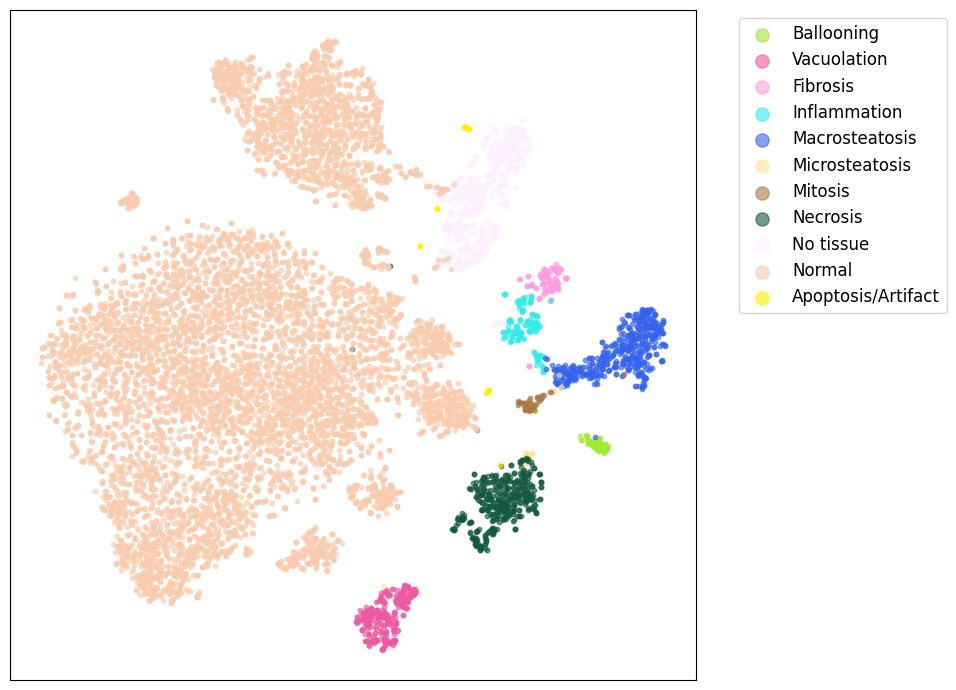}
\caption{t-SNE visualization of feature representations of pixels on the test set. Overall, t-SNE illustrates separability for major tissue patterns, while rare or visually ambiguous classes exhibit overlap. The unknown class (Apoptosis/Artifact) partially occupies a near-OOD region of the feature space illustrating challenging OOD detection.}
\label{fig:tSNE}
\end{figure}

\subsection*{A6. OOD detection methods}\label{sec:OOD_detection_methods}
\subsubsection*{Mahalanobis distance \textnormal{(Lee et al., 2018\cite{lee2018})}}

The underlying assumption of Mahalanobis distance-based OOD detection is that data points belonging to the same class exhibit similar feature representations which can be modeled as normally distributed. More precisely, feature representations $h(x)=\mathbf{h}$, 
conditioned on the ID-class label $y = c_i$,  $i\in\{0,..,K\}$, are assumed to follow a multivariate normal distribution,
\begin{center}
\(
p(\mathbf{h} \mid y = c_i) = \mathcal{N}(\mathbf{h}; {\boldsymbol{\mu}}_{c_i}, {\boldsymbol{\Sigma}})
\)
\end{center}
with a class-specific mean ${\boldsymbol{\mu}}_{c_i}$ and a pooled covariance matrix ${\boldsymbol{\Sigma}}$ which is shared across all classes, estimated as

\begin{center}
\(
\hat{\boldsymbol{\mu}}_{c_i} =
\displaystyle\frac{1}{N_{c_i}} \sum_{k : y_k = c_i} \mathbf{h}_k,
\qquad
\hat{\boldsymbol{\Sigma}} = \frac{1}{N} 
\displaystyle\sum_{i} \sum_{k : y_k = c_i} 
(\mathbf{h}_k - \hat{\boldsymbol{\mu}}_{c_i})(\mathbf{h}_k - \hat{\boldsymbol{\mu}}_{c_i})^{\top},
\)
\end{center}
where $N_{c_i}$ is the number of training samples in class $c_i$ and $N$ is the total number of training samples. 
After fitting class-specific distributions in feature space, the Mahalanobis distance between a test feature $\mathbf{h}$ and a class mean $\hat{\boldsymbol{\mu}}_{c_i}$ is calculated:
\begin{center}
\(
D^2(\mathbf{h},\hat{\boldsymbol{\mu}}_{c_i}) = (\mathbf{h} - \hat{\boldsymbol{\mu}}_{c_i})
\hat{\boldsymbol{\Sigma}}^{-1}
(\mathbf{h} - \hat{\boldsymbol{\mu}}_{c_i})^{\top}.
\)
\end{center}
In contrast to Euclidean distance, the Mahalanobis distance accounts for the covariance structure of the data. It adapts to anisotropic data distributions, penalizing deviations along directions of low variance more strongly. This property makes it particularly suitable for usage in high-dimensional spaces, such as the ViT feature space, where correlations between dimensions become significant.

The final anomaly score is defined in Lee et al.\cite{lee2018} as the negative of the smallest Mahalanobis distance between the test feature and the class means:
\begin{center}
\(
s_{Maha}(x) = - \displaystyle\min_{c_i}D(\mathbf{h},\hat{\boldsymbol{\mu}}_{c_i}),
\)
\end{center}
Consequently, larger Mahalanobis distances (and lower anomaly scores) indicate a higher likelihood of the point being an outlier with respect to the class distribution.

\subsubsection*{Modified Mahalanobis distance}

Instead of using the original features $\mathbf{h}$, we estimate the class means and covariance matrix based on $\ell_2$-normalized features:
\begin{center}
\(
\hat{\mathbf{h}} = \mathbf{h}/ \|\mathbf{h}\|_2.
\)
\end{center}
Test features are also $\ell_2$-normalized before computing their anomaly score. As discussed in Müller et al.\cite{mueller2025}, ViT-based feature norms may vary strongly and violate the Gaussian assumption underlying the Mahalanobis distance estimation. It was shown in Müller et al.\cite{mueller2025} that the simple $\ell_2$-normalization of the features mitigates this effectively and consequently improves OOD detection.

In order for the Mahalanobis-based OOD to be effectively used with our adaptive threshold selecton strategy, we define
\begin{center}
\(
s_{Maha+}(x) = -D(\hat{\mathbf{h}}, \hat{\boldsymbol{\mu}}_{c_i=\hat{y}}),
\)
\end{center}
where, instead of finding a minimal distance over all classes, we take the distance to the predicted class. The class-specific score set subject to thresholding is therefore built upon data points which got predicted as ${c_i}$ and whose anomaly scores are the Mahalanobis distances to class ${c_i}$. Thus, we avoid discrepancy between the point selection and its relevant Mahalanobis distance selection.

\end{document}